\documentclass{isprs}
\usepackage{subfigure}
\usepackage{setspace}
\usepackage{geometry}
\usepackage{epstopdf}
\usepackage[labelsep=period]{caption}
\usepackage[british]{babel} 
\usepackage[hang]{footmisc}

\usepackage[pagebackref=false,breaklinks=true,hidelinks,bookmarks=false]{hyperref}
\usepackage[all]{nowidow}
\usepackage{xspace}
\usepackage[table]{xcolor}
\usepackage{dirtytalk}
\usepackage{amsfonts}
\usepackage{amsmath}
\usepackage{amssymb}
\usepackage{url}
\usepackage{booktabs}
\usepackage{stfloats}
\usepackage{makecell}
\usepackage{array, makecell}
\usepackage{multirow}
\usepackage{enumitem}
\usepackage{adjustbox}
\usepackage{arydshln}
\usepackage{graphicx}
\usepackage{tabularx}
\usepackage[super]{nth}
\usepackage{comment}
\usepackage{float}
\usepackage[normalem]{ulem}  %

\xspaceaddexceptions{]\}}

\newcommand\mypara[1]{\vspace{1mm}\noindent\textbf{#1.}}

\newcommand{\ours}{\textsc{ImpliCity}\xspace}
\newcommand{\resdepth}{\textsc{ResDepth}\xspace}

\newcommand{\zurichOne}{\textsc{ZUR1}\xspace}
\newcommand{\zurichTwo}{\textsc{ZUR2}\xspace}
\newcommand{\zurichThree}{\textsc{ZUR3}\xspace}

\begin{document}

\title{\large\ours: City Modeling from Satellite Images with\\Deep Implicit Occupancy Fields\vspace{-2mm}}

\author{Corinne Stucker\thanks{Corresponding author} , Bingxin Ke, Yuanwen Yue, Shengyu Huang, Iro Armeni, Konrad Schindler\vspace{-3mm}}

\address{ETH Zurich, Switzerland -- (stuckerc, bingke, yuayue, shenhuan, iarmeni, schindler)@ethz.ch\vspace{-4mm}}

\icwg{}   %

\abstract{\vspace{-1em}High-resolution optical satellite sensors, combined with dense stereo algorithms, have made it possible to reconstruct 3D city models from space. However, these models are, in practice, rather noisy and tend to miss small geometric features that are clearly visible in the images. We argue that one reason for the limited quality may be a too early, heuristic reduction of the triangulated 3D point cloud to an explicit height field or surface mesh. To make full use of the point cloud and the underlying images, we introduce \ours, a neural representation of the 3D scene as an implicit, continuous occupancy field, driven by learned embeddings of the point cloud and a stereo pair of ortho-photos. We show that this representation enables the extraction of high-quality DSMs: with image resolution 0.5$\,$m, \ours reaches a median height error of $\approx\,$0.7$\,$m and outperforms competing methods, especially w.r.t.\ building reconstruction, featuring intricate roof details, smooth surfaces, and straight, regular outlines.\vspace{-1mm}}
\keywords{3D Reconstruction, Digital Surface Model (DSM), Deep Implicit Fields, Scene Representation, Satellite Imagery.\vspace{-1mm}}
\maketitle

\sloppy

\section{Introduction}
Modern very high-resolution (VHR) satellite sensors have made it possible to reconstruct sub-meter resolution 3D surface models from space. They are able to collect optical images with ground sampling distances $\leq$0.5$\,$m from multiple viewpoints almost anywhere on Earth.
Several software packages have been developed to derive 3D models from such satellite images \cite{krauss2013fully,de2014automatic,qin2016rpc,rupnik2017micmac,beyer2018ames,cournet2020ground,youssefi2020cars}. Typically, they adopt stereo matching algorithms originally developed for terrestrial or airborne photogrammetry. The principle of such algorithms is to find a dense set of image correspondences that have high photo-consistency and at the same time form a (piece-wise) smooth surface. After matching all suitable image pairs, the correspondences are triangulated to 3D points and fused into a single point cloud, which is commonly rasterized into a 2.5-dimensional height field (a.k.a.\ digital surface model, DSM) for further use.

\begin{figure}[!t]
\centering
\includegraphics[trim={0 2mm 0 2mm}, clip,width=\columnwidth]{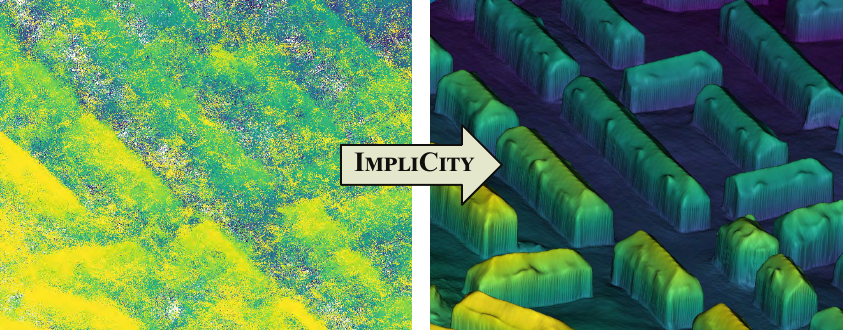}\vspace{-0.2em}
\caption{\ours is a deep, implicit representation of surface geometry. It is derived from a photogrammetric 3D point cloud and associated images. Note the geometric details on the roofs despite the substantial noise level of the input point cloud.}
\label{fig:teaser}
\end{figure}

Due to limited image resolution, sub-optimal stereo geometry, and radiometric differences caused by variable lighting and atmospheric effects, DSMs derived from satellite observations tend to be noisy (see Figure~\ref{fig:teaser}). Moreover, high-frequency details that would, in principle, be visible in the images are barely reconstructed. Those DSMs are thus often regarded as intermediate products and processed further, with a refinement step that aims to suppress noise and to impose a-priori assumptions about the surface, like straight building edges and vertical walls. Early attempts used low-level filtering and hand-coded rules. More recent works rely on neural networks to learn the mapping from a coarse DSM to a refined one from data~\cite{bittner2019dsm,bittner2019multi,bittner2020long,wang2021machine,stucker2022resdepth}. %

A fundamental property shared by different DSM reconstruction and refinement methods is an explicit representation of the surface, either as a mesh with a given number of vertices (respectively, faces) or as a regular 2D grid of height values. Such explicit parametrizations are convenient, but they do not preserve all information contained in the original point cloud and restrict the ability to resolve small structures. Recently, implicit neural functions have emerged as a powerful and effective representation of 3D geometry~\cite{park2019deepsdf,chen2019learning,mescheder2019occupancy,peng2020convolutional}. Instead of discretizing the 3D scene into a set of explicit surface elements, they implicitly model its geometry as a continuous field of occupancies or signed distance values, encoded in the weights of a neural network. The network can be evaluated at any 3D coordinate and, therefore, conceptually, allows for infinite resolution---in practice, its effective resolution is bounded by the representation power of the finite number of neurons, as well as by the resolution of the training data.

So far, implicit representations have been explored to model the 3D geometry of local shapes~\cite{genova2019learning,genova2020local}, single objects~\cite{park2019deepsdf,atzmon2020sal}, indoor scenes~\cite{jiang2020local,peng2020convolutional,sitzmann2020siren,chabra2020deep}, and single buildings~\cite{chen2021reconstructing}. In this work, we go one step further and investigate their potential to accurately reconstruct 3D urban scenes, on the order of several km\textsuperscript{2}, from satellite data. To that end, we introduce \ours, a coordinate-based, implicit neural 3D scene representation based on a point cloud derived from satellite photogrammetry. Since such point clouds are comparatively sparse and lack high-frequency detail, we additionally use an image stereo pair to guide the occupancy prediction. \ours reconstructs city models with fine-grained shape details, smooth and well-aligned surfaces, and crisp edges. It thereby reduces the mean absolute error by \textgreater60\% compared to a conventional stereo DSM.

\section{Related Work}
\mypara{Deep Implicit Functions}
Deep implicit functions for surface reconstruction have been proposed concurrently by \cite{mescheder2019occupancy,park2019deepsdf,chen2019learning}. These seminal works represent a 3D shape as an implicit, continuous field $f$, which is parametrized as a neural \emph{decoder} network, and constrained by a global latent \emph{code} (a \say{feature vector} of the scene) extracted with neural \emph{encoder} network. The field $f$ can be queried with a 3D location \mbox{$\mathbf{x} \in \mathbb{R}^3$} and returns either the \emph{occupancy} of $\mathbf{x}$ (i.e., its probability of lying below the surface) or its signed distance to the surface. To extract an explicit surface model, one reconstructs the iso-surface $f\!=\!0.5$ of the occupancy, respectively $f\!=\!0$ for the signed distance, for instance with marching cubes \cite{lorensen1987marching}.

As the scene information is stored as a global latent code, the method described so far does not generalize to unseen objects, fails to capture local surface details, and scales poorly with scene size. Therefore, more recent works \cite{chabra2020deep,jiang2020local,genova2020local,peng2020convolutional} decompose the scene into parts that are constrained by local codes. Moreover, \cite{peng2020convolutional} introduce a fully convolutional encoder. In this way, the implicit representation inherits the translation equivariance of convolutions; which, in turn, enables large-scale reconstructions. \cite{saito2019pifu} introduce local latent codes that are pixel-wise aligned with the image used for supervision, so as to obtain crisp surface edges aligned with the image gradients. The work perhaps most similar in spirit to ours is \cite{yang2021s3}, where an implicit neural model is used to reconstruct humans from LiDAR scans, guided by a (single) image to retrieve details such as the wrinkles of clothes.

\mypara{Deep Implicit Functions for Satellite Images}
To the best of our knowledge, \cite{derksen2021shadow,xiangli2021citynerf} are so far the only works that have explored deep implicit representations in the context of satellite data. Both are based on the Neural Radiance Field (NeRF) method of \cite{mildenhall2020nerf} that models an observed 3D scene as a continuous, volumetric field of viewpoint-dependent radiance values. The NeRF approach is designed primarily for novel-view synthesis, not geometrically detailed reconstruction. It does, of course, implicitly capture 3D geometry, but with an accuracy just enough to render it from new viewpoints and obtain radiometrically convincing images. E.g., the average reconstruction errors reported in \cite{derksen2021shadow} are similar to those of conventional satellite photogrammetry. Also, the NeRF encoder must capture viewpoint-dependent appearance changes, and therefore extract implicit lighting and material information. As a consequence, it cannot generalize beyond the training region. 

\section{Method}
\begin{figure*}[!ht]
\centering
\includegraphics[trim={0 0.7mm 0 1mm}, clip, scale=0.99]{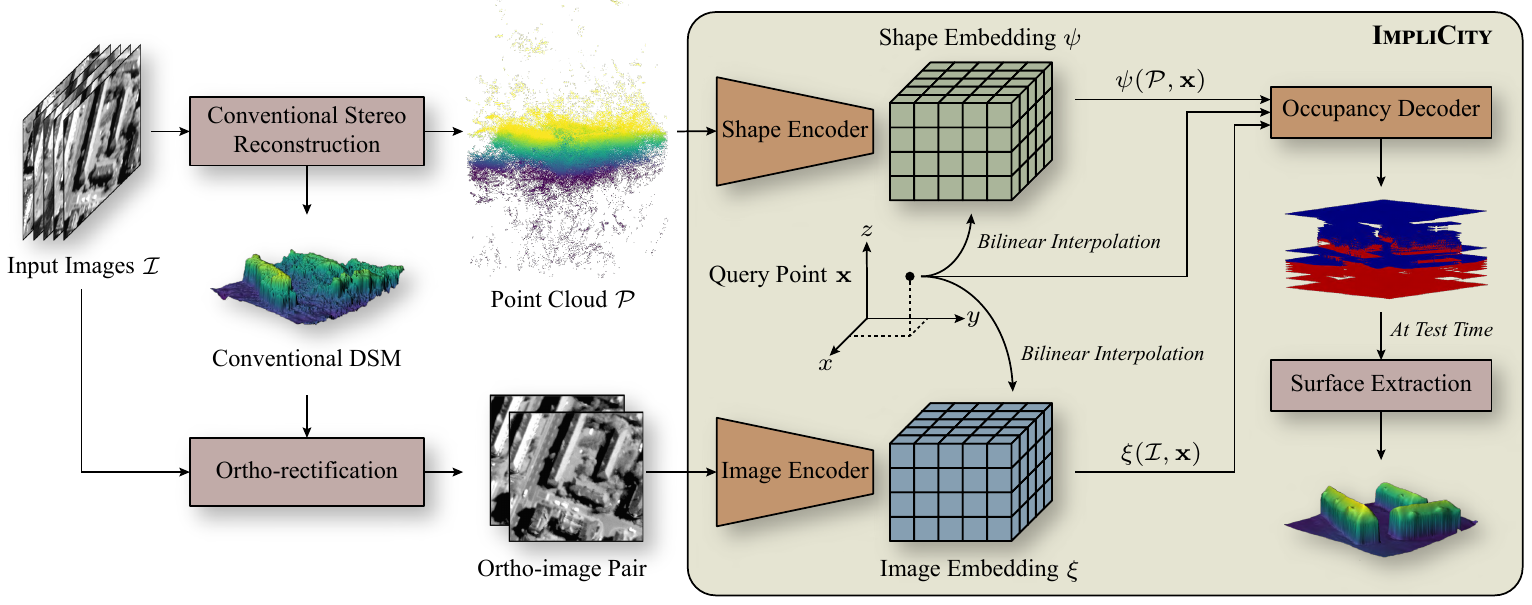}\vspace{-0.2em}
\caption{Method overview. Satellite images are processed into a 3D point cloud and a coarse DSM as a basis for ortho-rectification (left side). \ours takes the point cloud and ortho-photos as input and transforms them into a shape embedding $\psi$ and an image embedding $\xi$, which can be decoded into occupancy values in continuous 3D space to recover a high-accuracy DSM.}
\label{fig:overview}
\end{figure*}

Our approach starts from a set of satellite images $\mathcal{I}$ with overlapping fields of view and known camera poses. We follow best practices for satellite photogrammetry and first perform conventional, dense image matching for all suitable image pairs, followed by triangulation. See Section~\ref{sec:data}.

\mypara{Problem Formulation} Given a set of triangulated 3D points $\mathcal{P}=\left \{ \mathbf{p}_i \in \mathbb{R}^3 \right \}_{i=1}^N$, collected from all stereo pairs, our goal is to build a detailed and geometrically accurate 3D reconstruction of the observed scene. This is where our approach deviates from standard practice: we do not convert the raw point cloud~$\mathcal{P}$ into a raster DSM  for subsequent 2.5D processing. Instead, we reason in 3D space and represent the scene geometry as a continuous occupancy field. The field is represented by a function $f_\theta$ that, for a given 3D coordinate \mbox{$\mathbf{x} \in \mathbb{R}^3$}, returns the probability $\hat{o}$ that the location is occupied. I.e., $f_\theta$ should be 0 wherever there is free space, and 1 on and underneath the surface:
\begin{equation}
    f_{\theta}\big(\mathbf{x}, \psi(\mathcal{P}, \mathbf{x}), \xi(\mathcal{I}, \mathbf{x})\big) \rightarrow \hat{o} \in [0,1]\;,
\end{equation}
where $\psi(\mathcal{P}, \mathbf{x}) \in \mathbb{R}^d$ and $\xi(\mathcal{I}, \mathbf{x}) \in \mathbb{R}^d$ are location-dependent latent codes that modulate the occupancy probability. In our case, $\psi(\mathcal{P}, \mathbf{x})$ describes the local structure of the point cloud~$\mathcal{P}$, whereas $\xi(\mathcal{I}, \mathbf{x})$ encodes the local image texture at the 2D projections of $\mathbf{x}$ in two input views.
Inspired by recent trends in 3D surface reconstruction, we parametrize $f_\theta$ as well as the feature extractors $\psi(\mathcal{P}, \mathbf{x})$ and $\xi(\mathcal{I}, \mathbf{x})$ as neural networks. To extract an explicit surface from this implicit representation, one must sample a sufficiently dense set of 3D locations $\mathbf{x}$, evaluate the function $f_\theta$ at all of them, and extract the iso-surface $f_\theta=0.5$.

\mypara{Overview}
Figure~\ref{fig:overview} depicts an overview of our approach. At its core is \ours, a coordinate-based neural representation of 3D scene geometry, guided by satellite images. The inputs to \ours are a raw, irregular, and unoriented point cloud $\mathcal{P}$, as obtained from satellite-based stereo reconstruction, and two ortho-rectified (panchromatic) satellite images.
We first map every point in $\mathcal{P}$ to a feature vector that encodes its geometric context, then aggregate those feature vectors into a shape embedding~$\psi$. The embedding~$\psi$ is aligned with the geographic coordinates, i.e., its $z$-axis is the vertical and its $(x,y)$-axes are the East and North directions in the local UTM zone.
Similarly, we map the ortho-images to an image embedding~$\xi$ with a fully convolutional 2D encoder, so as to encourage consistency between the reprojected 3D scene geometry and the image content. Note that, since the ortho-images are rectified to UTM coordinates, the shape embedding~$\psi$ and the image embedding~$\xi$ are, by construction, aligned and share the same $(x,y)$-axes.
With these embeddings, we can, at any location $\mathbf{x}$, read out the two $d$-dimensional codes $\psi(\mathcal{P}, \mathbf{x}), \xi(\mathcal{I}, \mathbf{x})$ and pass them, together with the coordinates, to a decoder function \mbox{$f_{\theta}: \mathbb{R}^3 \times \mathbb{R}^d \times \mathbb{R}^d \rightarrow [0, 1]$} that infers the occupancy state at $\mathbf{x}$. In our case, the decoder is a multi-layer perceptron with internal skip connections.

In the following, we introduce our network architecture (Section~\ref{sec:architecture}) and its variants (Section~\ref{sec:network_variants}) in more detail before proceeding to the training procedure (Section~\ref{sec:training}) and the sampling strategy employed to define the training signal (Section~\ref{sec:sampling}). Finally, we describe how we convert the implicit occupancy volume into an explicit raster DSM (Section~\ref{sec:dsm_generation}).

\subsection{Network Architecture}
\label{sec:architecture}
The architecture of \ours builds upon recent advances in learned, implicit 3D modeling. We adopt the convolutional single-plane encoder proposed by \cite{peng2020convolutional} to process the input point cloud and the pixel-aligned encoder of \cite{saito2019pifu} to process the ortho-images. As a decoder, we apply the same fully-connected network as in \cite{peng2020convolutional}.

\mypara{Shape Embedding}
To represent the local 3D point distribution that forms the basis for surface reconstruction, we compute a feature encoding from the point cloud $\mathcal{P}$. We follow \cite{peng2020convolutional} and first apply a point-wise encoder based on PointNet~\cite{qi2017pointnet}, with one fully connected layer followed by five fully-connected ResNet blocks~\cite{he2016deep}. Each ResNet block includes a local pooling operation to locally aggregate 3D context information. The extracted $d$-dimensional per-point features are then orthographically projected onto a horizontal plane and discretized into a regular 2D grid of $H \times W$ grid cells, where features that project into the same cell are averaged. In our implementation, we use a grid spacing of 0.5$\,$m in world coordinates. Following \cite{peng2020convolutional}, the resulting feature \say{image}, with size $H \times W \times d$, is processed further with a 2D U-Net~\cite{ronneberger2015u}, equipped with symmetric skip connections to preserve high-frequency information. To capture long-range context, the depth of the U-Net is set such that its receptive field spans the entire feature image.

\mypara{Satellite Image Embedding}
Point clouds derived from satellite images are comparatively sparse and fairly noisy (cf. Figure~\ref{fig:teaser}). As a consequence, they do not preserve high-frequency details (like sharp roof edges or small dormers) that are, in principle, visible in the images. To recover fine-grained geometric details, we thus build a second latent embedding~$\xi$ from a panchromatic stereo pair. That image embedding is then used as additional input to the decoder to guide the occupancy prediction. The two images of the stereo pair are aligned by ortho-rectifying both of them with the same, preliminary surface model (cf.~Section~\ref{sec:data}) and stacked into a two-channel image. To generate $\xi$, we process that image with an encoder similar to the stacked hourglass architecture~\cite{newell2016stacked} used in PIFu~\cite{saito2019pifu}. To adapt it for our purposes, we modify the first layer to accept our two-channel input and change the hidden feature dimension $d$ to match that of the shape embedding~$\psi$. Note that ortho-rectifying the images  \textit{(i)}~makes it possible to work with a single image embedding despite the two different viewpoints, and \textit{(ii)}~ensures that the embeddings $\psi$ and $\xi$ are correctly aligned.

\mypara{Occupancy Decoder}
The task of the decoder is to estimate the occupancy probability at any location in scene space. Given a point \mbox{$\mathbf{x} \in \mathbb{R}^3$}, we project it onto the horizontal $(x,y)$ coordinate plane and retrieve its shape code $\psi(\mathcal{P}, \mathbf{x})$ and image code $\xi(\mathcal{I}, \mathbf{x})$ from the two embeddings with bilinear interpolation.
The occupancy at $\mathbf{x}$, as a function of its coordinates $\mathbf{x}$, shape code $\psi(\mathcal{P}, \mathbf{x})$, and image code $\xi(\mathcal{I}, \mathbf{x})$, is then predicted with a network consisting of five consecutive, fully-connected ResNet blocks. In our implementation, each ResNet block has $d$ neurons, and the sum $\psi(\mathcal{P}, \mathbf{x})+\xi(\mathcal{I}, \mathbf{x})$ of the two codes is added as side input to every block, as in \cite{peng2020convolutional}.

\subsection{Network Variants}
\label{sec:network_variants}
In our method, the stereo images are simply stacked and encoded independently of the point cloud. This raises the question whether a single image might be enough, and whether the use of images improves the reconstruction at all. To investigate these questions, we construct 
\pagebreak
two network variants that differ w.r.t.\ the number and combination of input modalities but are otherwise identical. In particular, we keep the network architecture fixed and train each variant using the same training settings and data samples. The network configuration based on stereo guidance is our default setting, referred to as \ours-stereo (or simply \ours if not stated otherwise). The first variant, \ours-mono, uses only a single ortho-image to generate the latent embedding~$\xi$. Therefore, it cannot exploit stereo information (in the form of misalignment between ortho-photos) and has to make do with image patterns and textures from a single image, with no redundancy. The second variant, \ours-0, has no access to image information. It learns the mapping from 3D points to occupancies constrained only by the shape embedding $\psi$, i.e., the local point distribution. Note that this configuration corresponds to the original \textit{Convolutional Occupancy Networks} proposed in \cite{peng2020convolutional}.

\subsection{Training and Inference}
\label{sec:training}
At training time, we randomly sample query points $\left \{ \mathbf{x}_i \in \mathbb{R}^3 \right \}$ within the volume of interest and in the vicinity of the true surface (see Section~\ref{sec:sampling}). The training is supervised by the binary cross-entropy loss $\mathcal{L}$ between the predicted occupancies~$\hat{o}$ and the true occupancies~$o$ at these points:
\begin{equation}
    \mathcal{L}(\hat{o}, o) = \sum_i \big(o_i \cdot \log(\hat{o}_i)
    + (1 - o_i) \cdot \log(1 - \hat{o}_i)\big)\;.
    \label{eq:loss}
\end{equation}
True occupancies~$o_i$ are derived from an existing city model of the training region. At inference time, we sample a regular 3D grid of query points in a hierarchical fashion, see Section~\ref{sec:dsm_generation}. 

\subsection{Spatial Sampling}
\label{sec:sampling}
One challenge when training implicit neural shape models is to reach the right balance between expressiveness and generality, which boils down to sampling adequate 3D points~$\mathbf{x}_i$ during training. If points were uniformly sampled in 3D space, most points would be far away from all surfaces. Consequently, the learned model would be biased towards predicting free space, as the dominant class in the absence of strong surface cues; and towards overly smooth reconstructions, since it has rarely seen surface details during training. On the other hand, if the points were exclusively sampled in the vicinity of the surface, the model would be prone to overfitting the training set, since the learning would narrowly focus on specific properties of the training area that may not generalize to other parts of the space.

In our approach, we combine uniform sampling and surface sampling, a strategy that has proven efficient for implicit neural models~\cite{saito2019pifu}. To begin with, we uniformly sample a first set of points arbitrarily within the volume of interest. Second, we densely sample a second set of points on the true surface and perturb them with zero-mean Gaussian random noise, for our data with standard deviation $\sigma\!=\!0.4\,$m. See Section~\ref{sec:data} for details. The two sets are then merged and together form the training set. In our experiments, the ratio between arbitrary points and surface points is 1:4.

\subsection{Surface Extraction}
\label{sec:dsm_generation}
To turn the implicit function $f_\theta$ into an explicit surface representation, we use a conventional raster DSM with a grid spacing of 0.25$\,$m. Inspired by the \textit{Multi-resolution Iso-Surface Extraction} algorithm of \cite{mescheder2019occupancy}, we employ a hierarchical refinement scheme to extract the iso-surface $\hat{o}\!=\!0.5$ from the occupancy volume. This approach makes it possible to recover a high-resolution DSM without having to densely sample the entire height range.

We start by discretizing the volume of interest into a regular grid of 3D points with a horizontal resolution equal to the grid spacing of the DSM and an initial vertical resolution of 16$\,$m. Next, the occupancy of every grid point is predicted with the trained \ours model. Using the fact that in a 2.5D DSM there is exactly one transition per pixel from free to occupied space, we mark the highest occupied 3D point per $(x,y)$-column and the one immediately above it as active, increase the vertical resolution between the two active points by a factor 4, and predict the occupancy of the three newly generated points. Then, we again zoom in on the highest occupied point and the one immediately above it and repeat the refinement. Four iterations of this refinement lead to a final nominal resolution of 6.25$\,$cm in the vertical direction. The highest occupied point after the last iteration is declared the DSM height $z(x,y)$. Going down to such a low nominal resolution helps to avoid aliasing artefacts on the reconstructed surface, even though it is, of course, far below the effective vertical resolution achievable with satellite images of $\approx\,$0.5$\,$m GSD at nadir.

\section{Experiments}
\subsection{Dataset and Preprocessing}
\label{sec:data}

\mypara{Imagery and Study Area} We evaluate our method on panchromatic satellite images acquired over Zurich, Switzerland. We have one WorldView-3 and 14~WorldView-2 images at our disposal. They were captured between 2014 and 2018, with 22~days the shortest time interval between two acquisitions. The average GSD is $\approx\,$0.5$\,$m at nadir. The study area%
\footnote{The area corresponds to \textsc{ZUR1} of~\cite{stucker2022resdepth}.} %
covers 4{$\,$km\textsuperscript{2}} and includes widely spaced, detached residential buildings, allotments, and high commercial buildings. Moreover, it contains a stretch of the river Limmat and a forested hill. In analogy to~\cite{stucker2022resdepth}, we split the area into five equally large, mutually exclusive stripes and allocate three stripes for training, one for validation, and one for testing.

\mypara{Point Cloud Generation}
We use a re-implementation of state-of-the-art hierarchical semi-global matching~\cite{rothermel2012sure}, tailored to satellite images, to generate the input point cloud~$\mathcal{P}$. First, we employ the method of~\cite{patil2019new} to perform bias correction of the supplied rational polynomial coefficient (RPC) projection models. 
\pagebreak
Next, we determine suitable image pairs for dense matching based on heuristics inspired by \cite{facciolo2017automatic,qin2019critical}. Starting from all possible image pairs, we eliminate those whose intersection angles in object space are \textless5$^\circ$ or \textgreater30$^\circ$ (measured at the center of the region of interest), or whose incidence angles are \textgreater40$^\circ$ (mean of the two images). We further discard image pairs whose difference in sun angle is \textgreater35$^\circ$. To leverage the redundancy in the image set as much as possible, we use all remaining image pairs for pairwise rectification and pairwise dense matching, irrespective of differences in acquisition time, as suggested by \cite{krauss2019cross}. After matching, we use the inverse RPC projection function to triangulate corresponding points per image pair, resulting in 26~stereo clouds in the same scene coordinate system, which we simply merge into a single point cloud~$\mathcal{P}$.

\mypara{Initial DSM Reconstruction}
Besides the point cloud~$\mathcal{P}$, \ours receives two ortho-rectified panchromatic satellite images as input. For the ortho-rectification, we require an initial surface estimate of the observed scene. To generate it, we fuse the point cloud~$\mathcal{P}$ into a coherent multi-view raster DSM with a grid spacing of 0.25$\,$m, by computing the cell-wise median of the $n$ highest 3D points, where $n$ is defined as the average number of 3D points per grid cell~\cite{rothermel2016}. Further, we adopt standard post-processing operations from aerial and satellite-based photogrammetry
to denoise the DSM, remove spikes, and fill cells without a valid height with inverse distance weighted (IDW) interpolation.

\mypara{Stereo Pair Selection and Rectification}
Among all available image pairs with adequate stereo geometry (see above), we determine a single best pair that serves as the second input to \mbox{\ours}. The selection is based on three criteria, namely low intersection angle, small time difference between acquisitions (similar season), and low cloud coverage. Like~\cite{stucker2022resdepth}, we ortho-rectify the two selected images with the help of the initial DSM, without ray-casting to detect occlusions. Instead, duplicate gray-values are rendered for rays that intersect the surface twice, leading to systematic patterns of repeated, photometrically inconsistent textures. Due to the small baseline between the two views, discrepancies between the ortho-images (except for illumination and atmospheric effects) primarily stem from height errors in the initial DSM rather than from viewpoint differences.

\mypara{Ground Truth Occupancy}
To train our method, we need to know the true occupancy of any 3D spatial location sampled within the volume of interest. Fortunately, such full 3D supervision can be readily derived from the publicly available city model of Zurich~\cite{zurich-model}. The model has been created by the municipal surveying department in a semi-automatic manner, by fusing airborne laser scans, building and road boundaries (including bridges) from national mapping data, and roof models derived by manual stereo digitization. The height accuracy is specified as $\pm$0.2$\,$m for buildings and $\pm$0.4$\,$m for terrain. 

We densely sample points on roofs, facades, and terrain of the city model. The average distance between nearest points amounts to 0.2$\,$m for points sampled on facades and terrain. For points sampled on roofs, we increase the sampling resolution to 0.1$\,$m to capture geometric details such as dormers with higher fidelity. Furthermore, we uniformly sample points within the volume of interest with a mean distance of $\approx\,$1.0$\,$m between points. Points sampled on the surface are assigned true occupancy values of 1; points sampled in free space are assigned 0 or 1, depending on whether they lie above or below the surface.

\subsection{Implementation Details}
We randomly sample training patches with a spatial dimension of 64$\times$64$\,$m in world coordinates from the training region. To avoid biases due to the specific topography and urban layout, we augment the data by randomly rotating the training patches by \mbox{$\alpha\in\{0^\circ,90^\circ,180^\circ,270^\circ\}$} and random flipping along the $x$ and $y$ axes. At inference time, we reconstruct large-scale scenes by applying the learned model in a sliding window.

We follow best practice and normalize the data (point cloud, ortho-images, query points) for neural network training. Every 64$\times$64$\,$m patch, originally given in UTM coordinates (zone 32T), is first horizontally shifted and scaled such that all point coordinates lie in $[0,1]$. Then, all points are vertically centered to the median height and rescaled with a fixed factor. That factor is found by computing standard deviations of the heights for 20'000 random patches from the training set and averaging them (cropped to the \nth{5} and \nth{95} percentile for robustness). Ortho-images are normalized with the mean and standard deviation over the intensity values of all training pixels.

We have implemented \ours in PyTorch and run it on a NVIDIA GeForce GTX 2080 Ti GPU. Source code and pretrained models are available at \url{https://github.com/prs-eth/ImpliCity}. In all experiments, we use a hidden feature dimension~$d$ of 32 for both encoders and the joint decoder, and feature plane dimensions 128$\times$128 for the shape embedding~$\psi$ and 64$\times$64 for the image embedding~$\xi$. For training, we employ the \textsc{ADAM} optimizer with a base learning rate of \mbox{$5 \cdot 10^{-5}$} ($\beta_1$=0.9, $\beta_2$=0.999), no weight decay, and a cyclical learning rate scheduler~\cite{smith2017cyclical} with cycle amplitude \mbox{$5 \cdot 10^{-4}$}. We set the batch size to 1 but accumulate gradients for 64~training iterations before performing back-propagation. Errors at water and forest pixels are down-weighted by a factor of 0.5 when computing the loss (Eq.~\ref{eq:loss}). We stop training once the DSM metrics (cf.\ Section~\ref{sec:metrics}) on the validation set have converged. We experimentally found that reconstruction quality improves when areas with evident temporal differences between the satellite imagery and the city model are masked out during training.

\subsection{Baselines}
\label{sec:baselines}
We compare \ours against the following baselines:

\textbf{Initial DSM}: The raster DSM generated from the input point cloud~$\mathcal{P}$, representative of conventional satellite-based reconstruction (see Section~\ref{sec:data} for details).

\textbf{\resdepth}: A learned DSM refinement approach by \cite{stucker2022resdepth} that directly refines the initial raster DSM with a U-Net~\cite{ronneberger2015u}. \resdepth-0 is trained to regress an additive height correction at every pixel. The image-guided variants \resdepth-mono and \resdepth-stereo exploit one and two ortho-images as additional input to guide the refinement.

\textbf{PIFu}: The Pixel-aligned Implicit Function (PIFu) method of \cite{saito2019pifu}, representative for deep, implicit surface reconstruction from images. We feed the initial DSM as input to the network. This baseline, denoted \mbox{PIFu-0}, corresponds to learned DSM filtering with the help of an implicit neural scene representation. Moreover, we train PIFu-mono and PIFu-stereo variants with one, respectively two ortho-images as additional input channels. 
To remain consistent with the other methods, we use a patch size of 256$\times$256 pixels (64$\times$64$\,$m in scene space) rather than 512$\times$512 as in the original PIFu.

\subsection{Quality Metrics}
\label{sec:metrics}
We use the publicly available city model of Zurich~\cite{zurich-model} to evaluate the performance of \ours. That city model is delivered in the form of 2.5D building models and a terrain surface. Therefore, we resort to 2.5D metrics commonly used for DSM evaluation.
Regions where the city model differs from the images due to recent construction activities have been masked out.
We render a reference DSM from the city model to measure the mean absolute error (MAE), the root mean square error (RMSE), and the median absolute error (MedAE), computed over per-pixel deviations between predicted and reference heights.%
\footnote{These widely used pixel-wise metrics do not fully characterize DSM quality: improved  reconstruction of intricate geometric details may not be reflected in lower errors, see Sec.~\ref{subsec:results}.} %
For a more in-depth analysis, we calculate the metrics separately for building and terrain pixels (according to the ground truth), where the building mask has been dilated by two pixels (0.5$\,$m) to ensure that distortions along its contours are reflected in the building error. Moreover, we differentiate between general terrain and forested areas with the help of a manually created forest mask.

\subsection{Results}
\label{subsec:results}

\begin{table*}[!ht]
    \setlength\dashlinedash{2pt}
    \setlength\dashlinegap{1.5pt}
    \setlength\arrayrulewidth{0.3pt}
    \centering
	\begin{adjustbox}{max width=0.95\textwidth}
        \begin{tabular}{@{\hspace{\tabcolsep}}lcccccccccccc@{\hspace{\tabcolsep}}}
    	    \toprule
    		Reconstruction &
    		\multicolumn{3}{c}{Overall} & \multicolumn{3}{c}{Buildings} & \multicolumn{3}{c}{Terrain} & \multicolumn{3}{c}{Terrain w/o forested areas}\\
    		\cmidrule(lr){2-4}\cmidrule(lr){5-7}\cmidrule(lr){8-10}\cmidrule(lr){11-13} 
    		& MAE & RMSE & MedAE & MAE & RMSE & MedAE & MAE & RMSE & MedAE & MAE & RMSE & MedAE \\
    		\midrule
    		Conventional DSM         & 	 3.89 & 7.03 & 1.59 & 3.02 & 5.02 & 1.47 & 4.29 & 7.78 & 1.65 &     2.79 &  4.69 &  1.43 \\
    		\addlinespace[0.5ex]
            \hdashline\noalign{\vskip 1ex}
    	    \resdepth-0         &    2.21 &  4.20 &  0.98 &     2.84 &  5.34 &  1.10 &    1.93 &  3.56 &  0.94 &     \cellcolor{lightgray}{1.50} &  \cellcolor{lightgray}{2.71} &  \cellcolor{lightgray}{0.84}\\
    	    PIFu-0              &    1.99 &  3.83 &  0.98 &     2.88 &  5.30 &  1.24 &    \cellcolor{lightgray}{1.58} &  \cellcolor{lightgray}{2.92} &  \cellcolor{lightgray}{0.89} &                \cellcolor{lightgray}{1.50} &  2.90 &  \cellcolor{lightgray}{0.84} \\
    	    \ours-0 (ours)     &    \cellcolor{lightgray}{1.87} &  \cellcolor{lightgray}{3.57} &  \cellcolor{lightgray}{0.92} &     \cellcolor{lightgray}{2.26} &  \cellcolor{lightgray}{4.55} &  \cellcolor{lightgray}{0.94} &    1.69 &  3.02 &  0.91 &     1.59 &  2.94 &  0.86 \\
    	    \addlinespace[0.5ex]
            \hdashline\noalign{\vskip 1ex}
            \resdepth-mono      &    1.65 &  3.22 &  0.77 &     2.07 &  4.18 &  0.91 &    1.45 &  2.67 &  0.72 &     1.20 &  2.27 &  0.65\\
            PIFu-mono           &    1.61 &  3.19 &  0.77 &     2.25 &  4.41 &  1.01 &    \cellcolor{lightgray}{1.32} &  \cellcolor{lightgray}{2.43} &  \cellcolor{lightgray}{0.69} &                \cellcolor{lightgray}{1.18} &  2.28 &  0.63 \\
            \ours-mono (ours)   &    \cellcolor{lightgray}{1.58} &  \cellcolor{lightgray}{3.03} &  \cellcolor{lightgray}{0.73} &     \cellcolor{lightgray}{2.00} &  \cellcolor{lightgray}{4.03} &  \cellcolor{lightgray}{0.80} &    1.39 &  \cellcolor{lightgray}{2.43} &  \cellcolor{lightgray}{0.69} &     1.23 &  \cellcolor{lightgray}{2.24} &  \cellcolor{lightgray}{0.62}\\
            \addlinespace[0.5ex]
            \hdashline\noalign{\vskip 1ex}
            \resdepth-stereo    &    1.53 &  2.97 &  0.74 &     \cellcolor{yellow}{1.91} &  3.93 &  0.82 &    1.35 &  2.41 &  0.71 &     1.15 &  2.12 &  0.65\\
            PIFu-stereo         &    1.53 &  3.04 &  0.73 &     2.13 &  4.28 &  0.91 &    \cellcolor{yellow}{1.20} &  \cellcolor{yellow}{2.20} &  \cellcolor{yellow}{0.66} &  \cellcolor{yellow}{1.11} &  \cellcolor{yellow}{2.11} &  0.61 \\
            \ours-stereo (ours) &       \cellcolor{yellow}{1.52} &  \cellcolor{yellow}{2.91} &  \cellcolor{yellow}{0.70} &     1.93 & \cellcolor{yellow}{3.86} &  \cellcolor{yellow}{0.78} &    1.33 &  2.35 &  0.67 &     1.20 &  2.22 &  \cellcolor{yellow}{0.60}\\
    	    \bottomrule
    	\end{tabular}
    \end{adjustbox}
    \vspace{-0.25em}
    \caption{Quantitative comparison of \ours with conventional DSM generation, and with learned DSM refinement based on either explicit (\resdepth) or implicit (PIFu) 2D representations (cf. Section~\ref{sec:baselines}). Methods are grouped according to their inputs: \textit{-0:}~point cloud/DSM, \textit{-mono}: point cloud/DSM and one ortho-image, \textit{-stereo}: point cloud/DSM and two ortho-images. All values are meters. {\setlength{\fboxsep}{1pt}\colorbox{lightgray}{Gray}} numbers are best for a given input, {\setlength{\fboxsep}{1pt}\colorbox{yellow}{yellow}} numbers are best overall.}
    \vspace{0.25em}
    \label{tab:ablation}
\end{table*}
\begin{figure*}[!ht]
    \def\mywidth{0.276\textwidth}  
    \setlength{\tabcolsep}{0.2em}
    \centering
    \begin{tabular}{l l m{\mywidth} m{\mywidth}  m{\mywidth}}
        \rotatebox[origin=c]{90}{\footnotesize Conventional DSM} &  &
        \includegraphics[width=\mywidth,trim={0 0 0 0},clip]{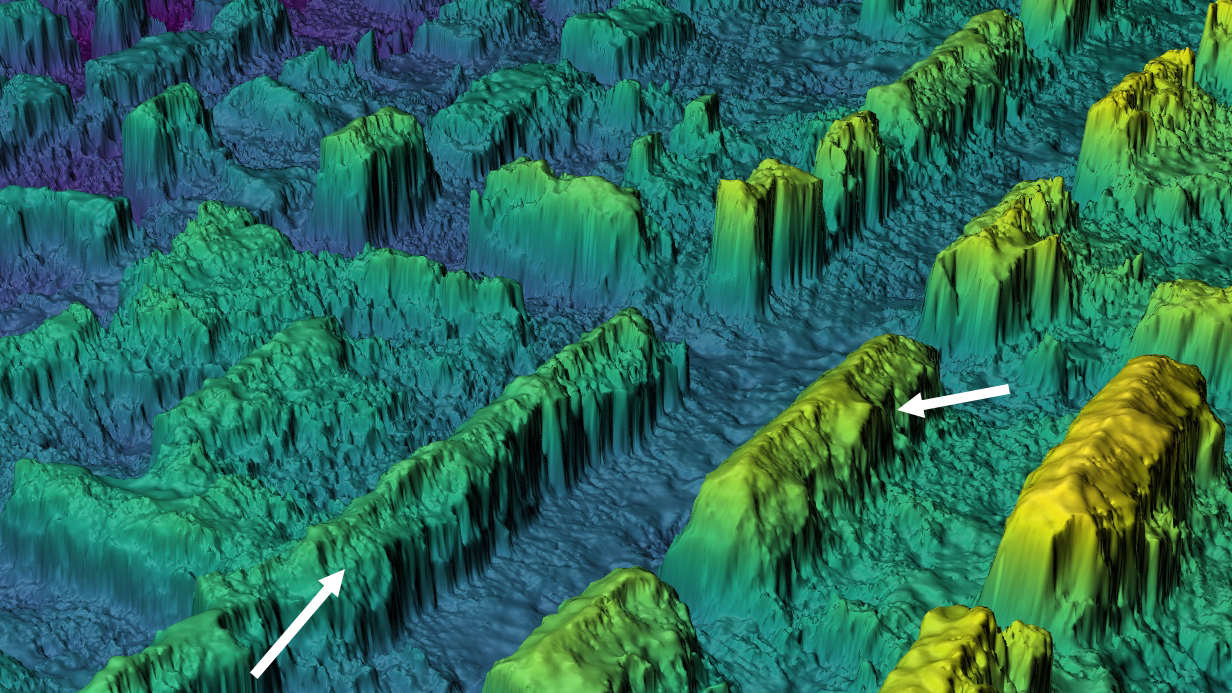} &
        \includegraphics[width=\mywidth,trim={0 0 0 0},clip]{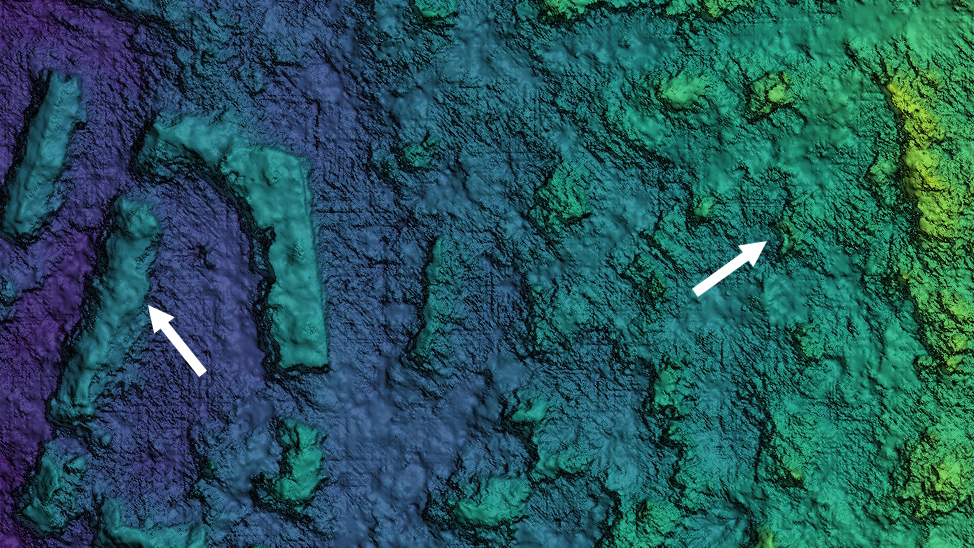} &
        \includegraphics[width=\mywidth,trim={0 0 0 0},clip]{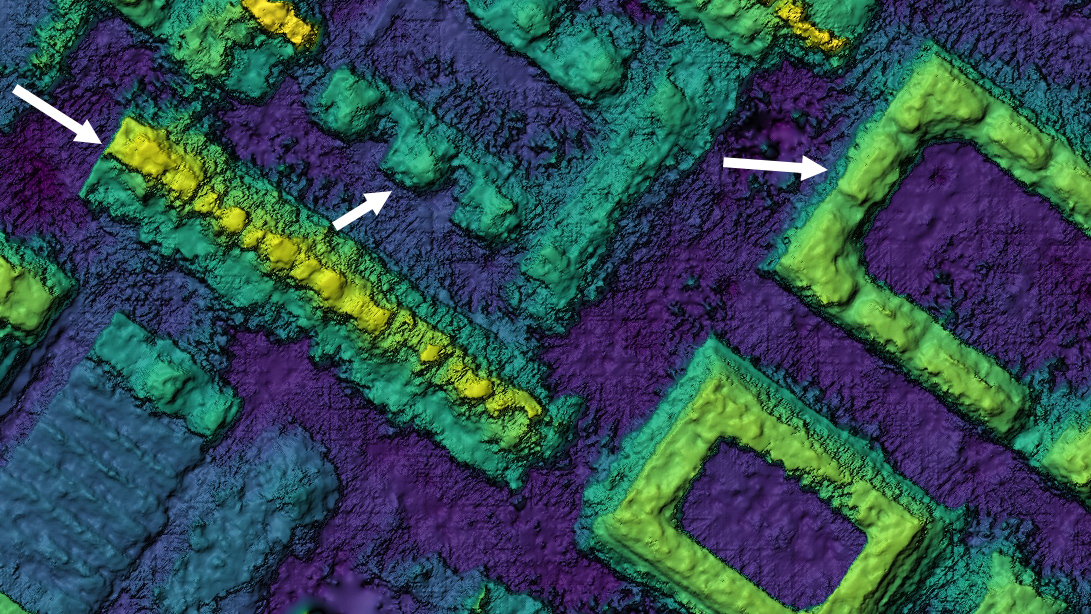}
        \\
        \rotatebox[origin=c]{90}{\footnotesize \resdepth-0} &  &
        \includegraphics[width=\mywidth,trim={0 0 0 0},clip]{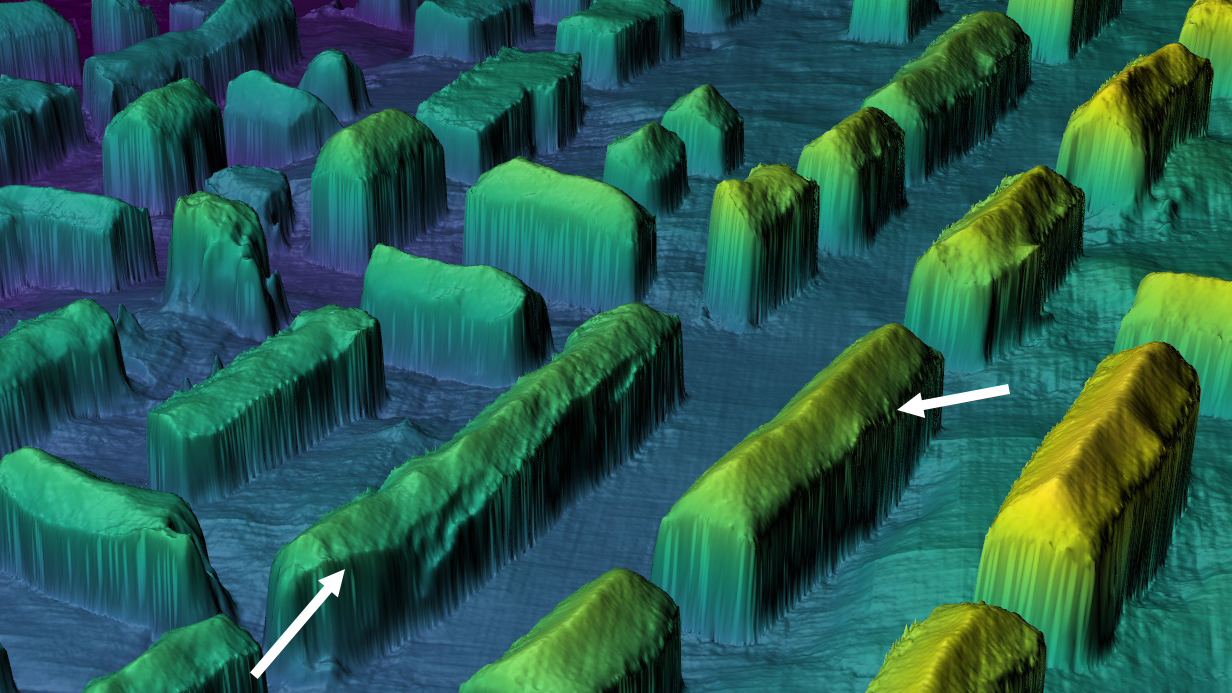} &
        \includegraphics[width=\mywidth,trim={0 0 0 0},clip]{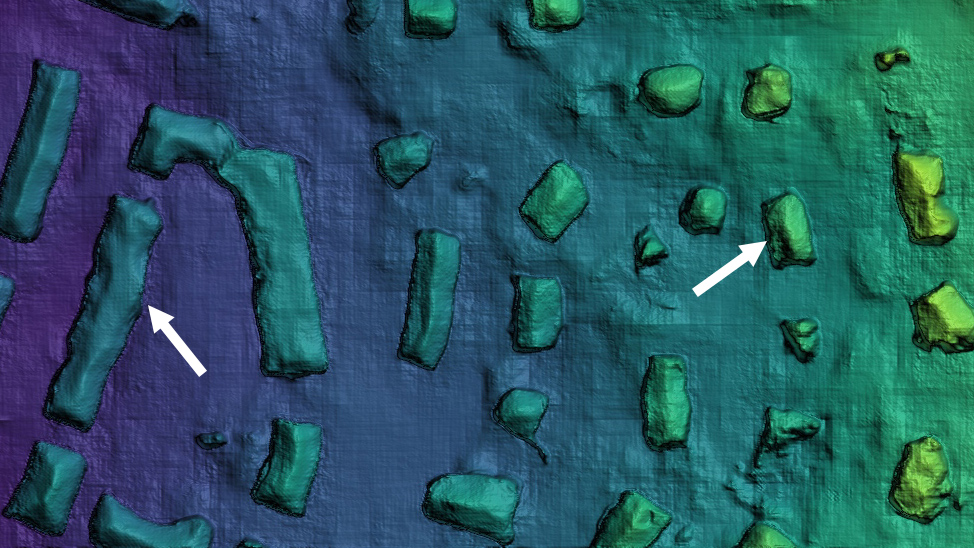} &
        \includegraphics[width=\mywidth,trim={0 0 0 0},clip]{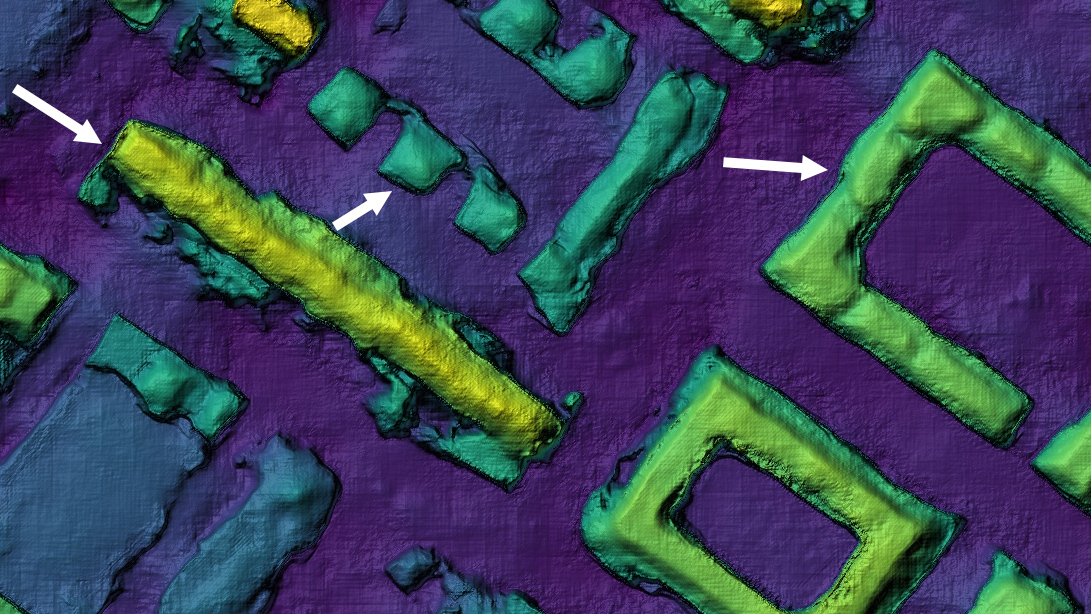}
        \\
        \rotatebox[origin=c]{90}{\footnotesize \resdepth-stereo} &  &
        \includegraphics[width=\mywidth,trim={0 0 0 0},clip]{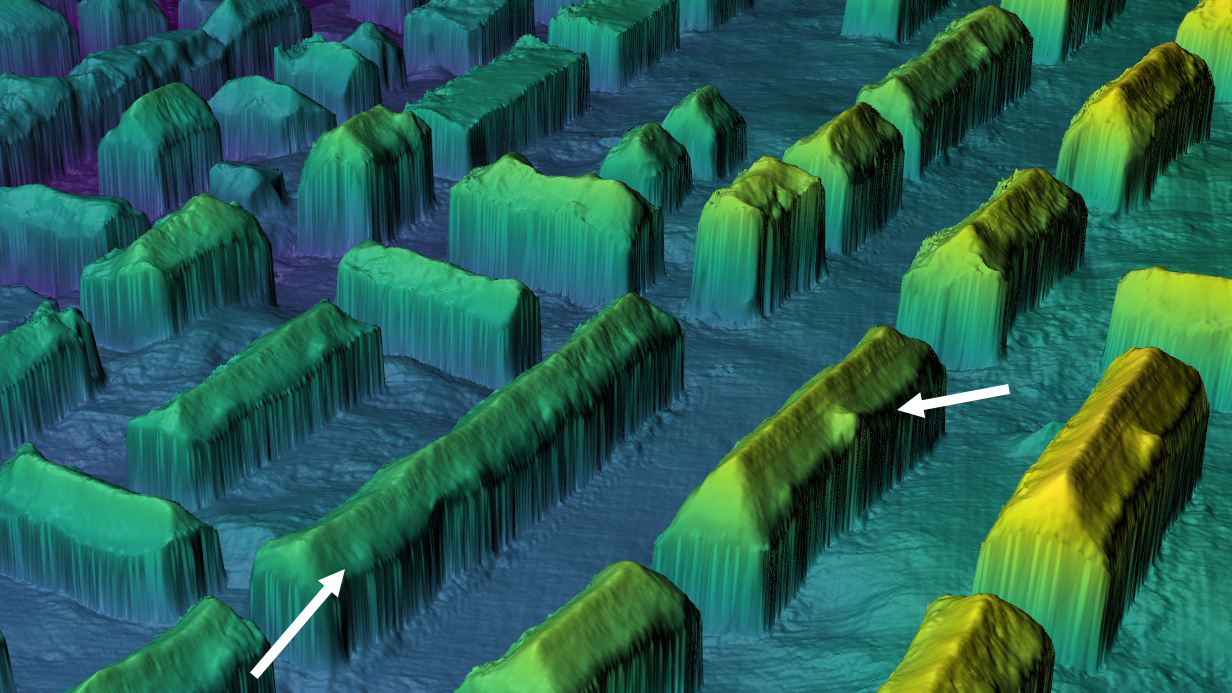} &
        \includegraphics[width=\mywidth,trim={0 0 0 0},clip]{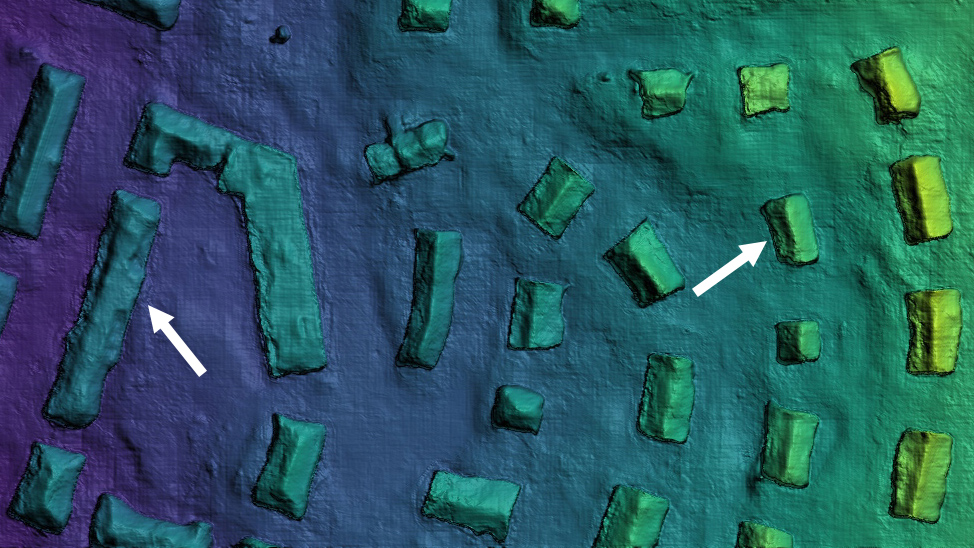} &
        \includegraphics[width=\mywidth,trim={0 0 0 0},clip]{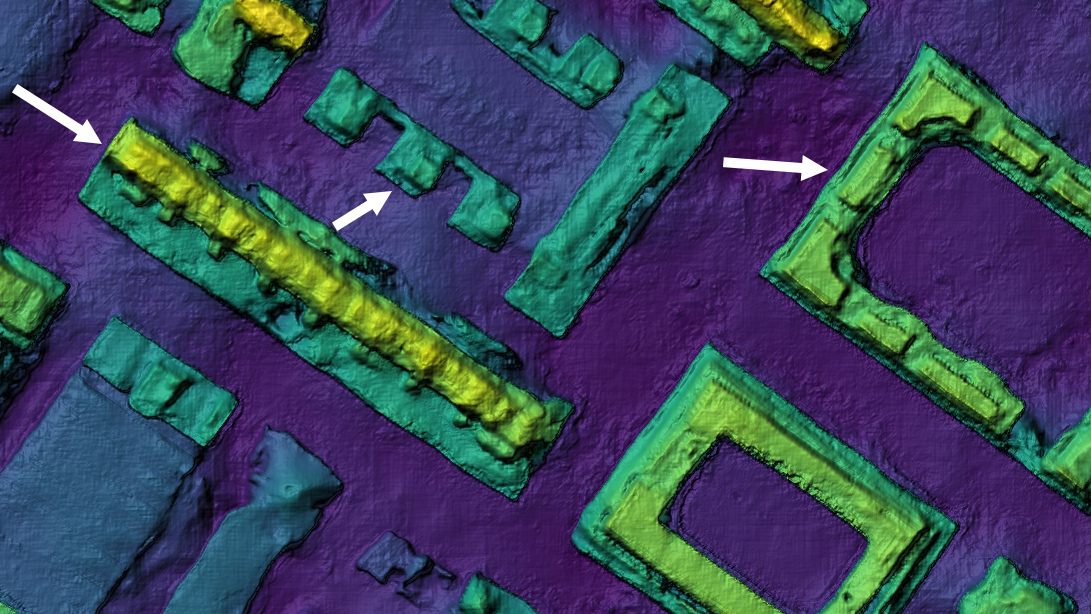}
        \\
        \rotatebox[origin=c]{90}{\footnotesize PIFu-stereo} &  &
        \includegraphics[width=\mywidth,trim={0 0 0 0},clip]{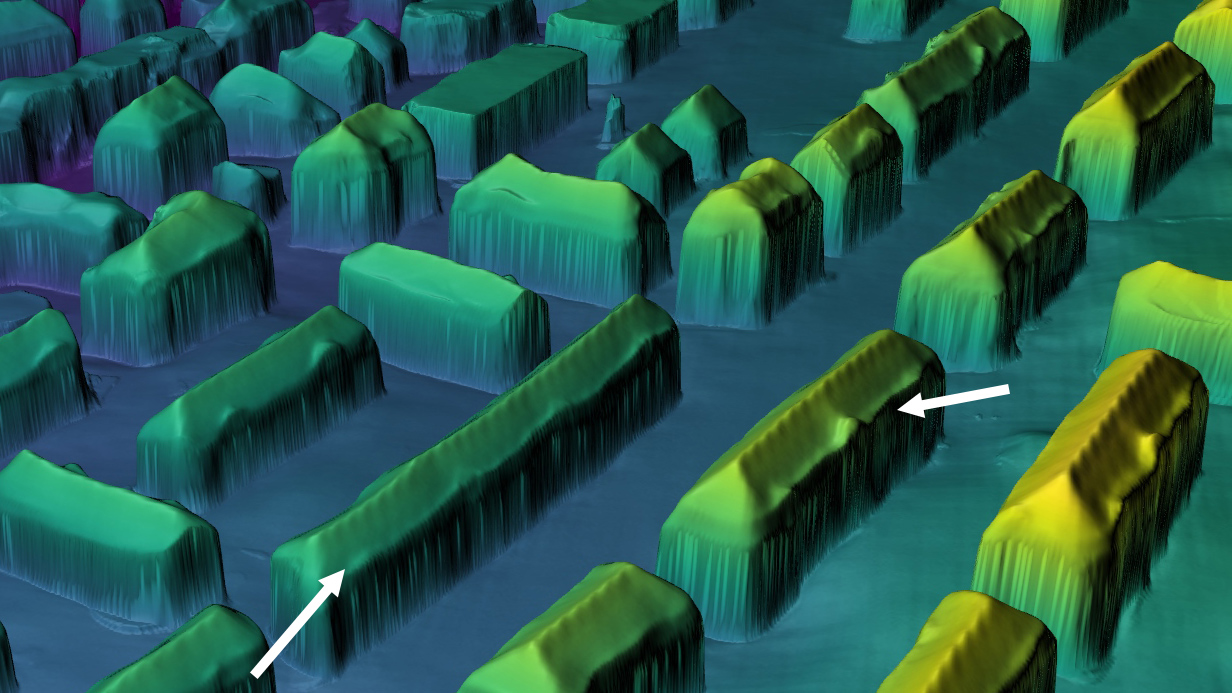} &
        \includegraphics[width=\mywidth,trim={0 0 0 0},clip]{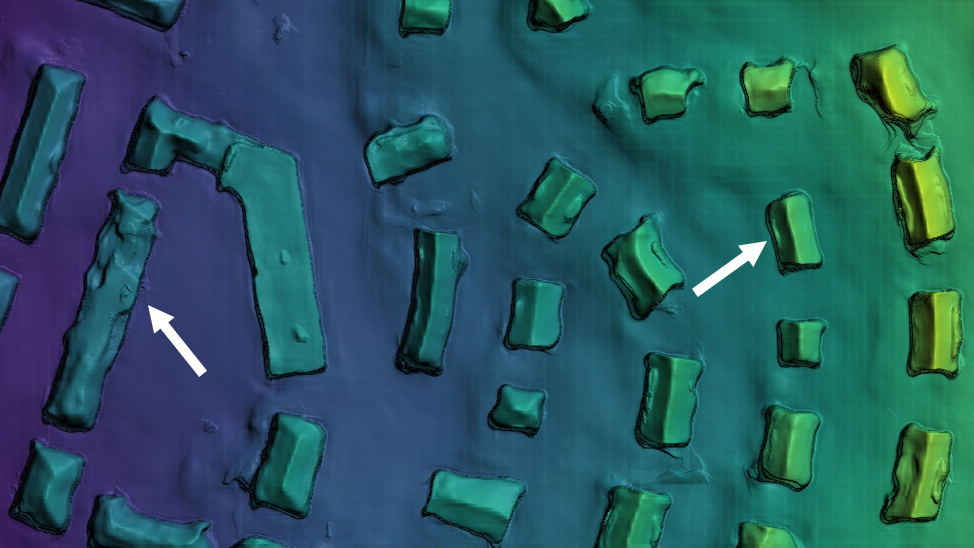} &
        \includegraphics[width=\mywidth,trim={0 0 0 0},clip]{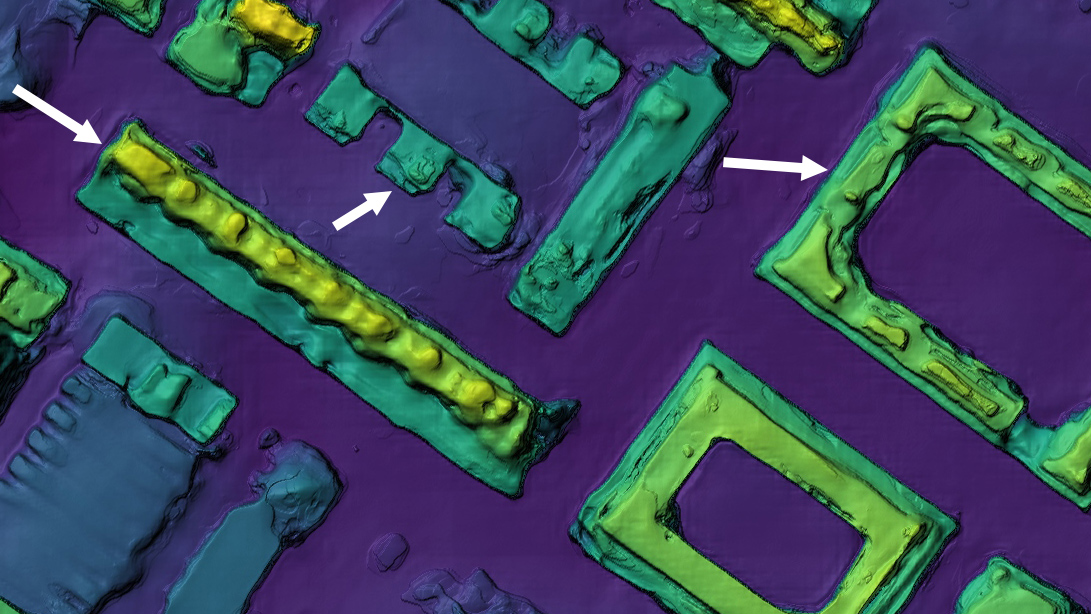}
        \\
        \rotatebox[origin=c]{90}{\footnotesize \ours-0} &
        \rotatebox[origin=c]{90}{\footnotesize (ours)} &
        \includegraphics[width=\mywidth,trim={0 0 0 0},clip]{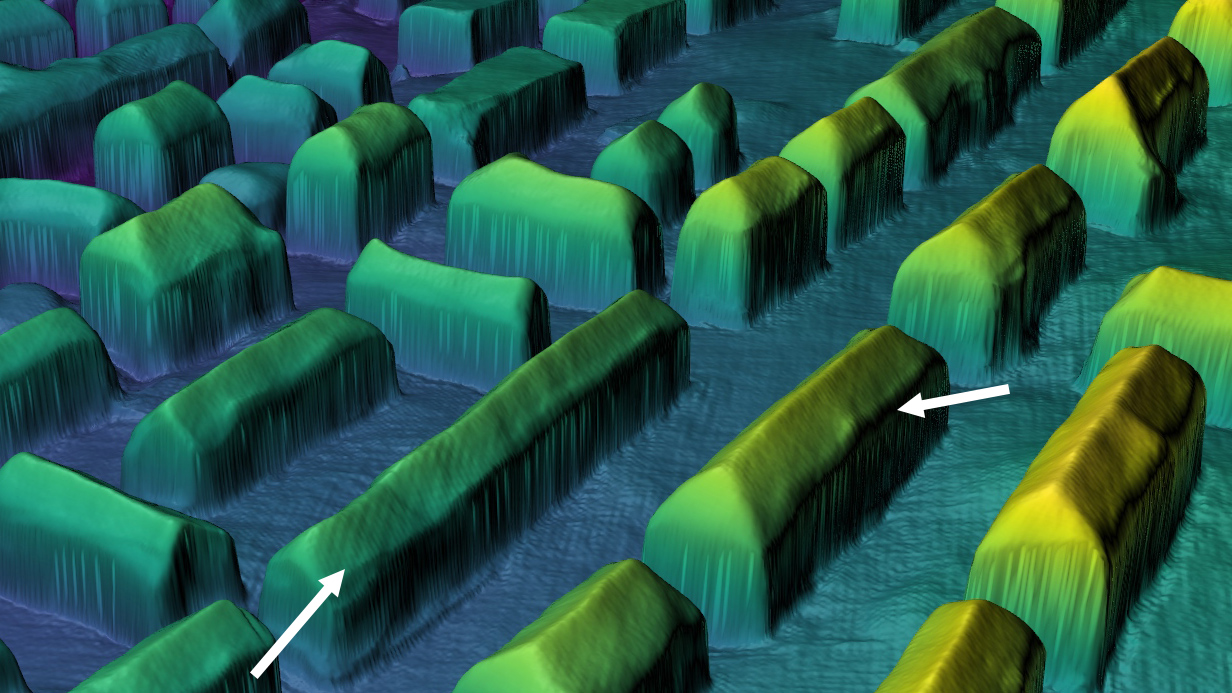} &
        \includegraphics[width=\mywidth,trim={0 0 0 0},clip]{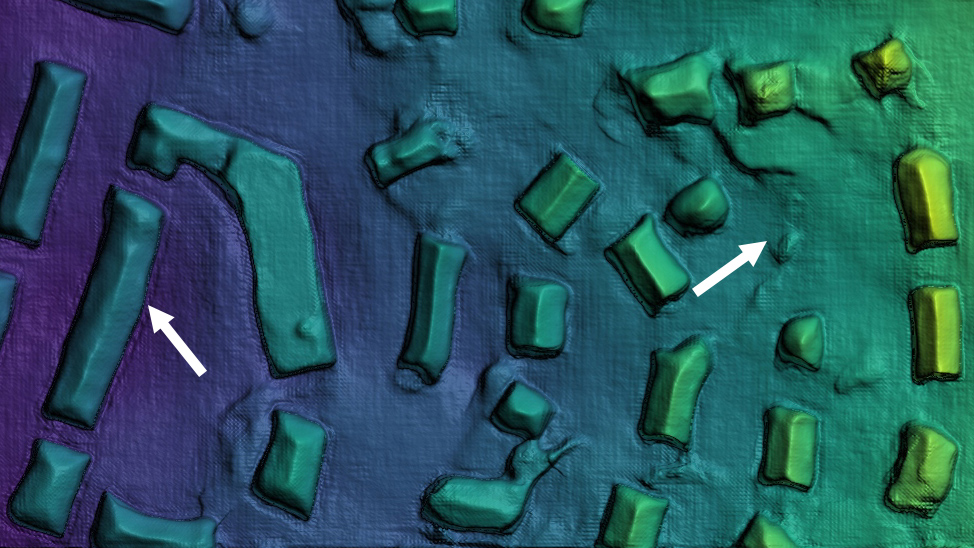} &
        \includegraphics[width=\mywidth,trim={0 0 0 0},clip]{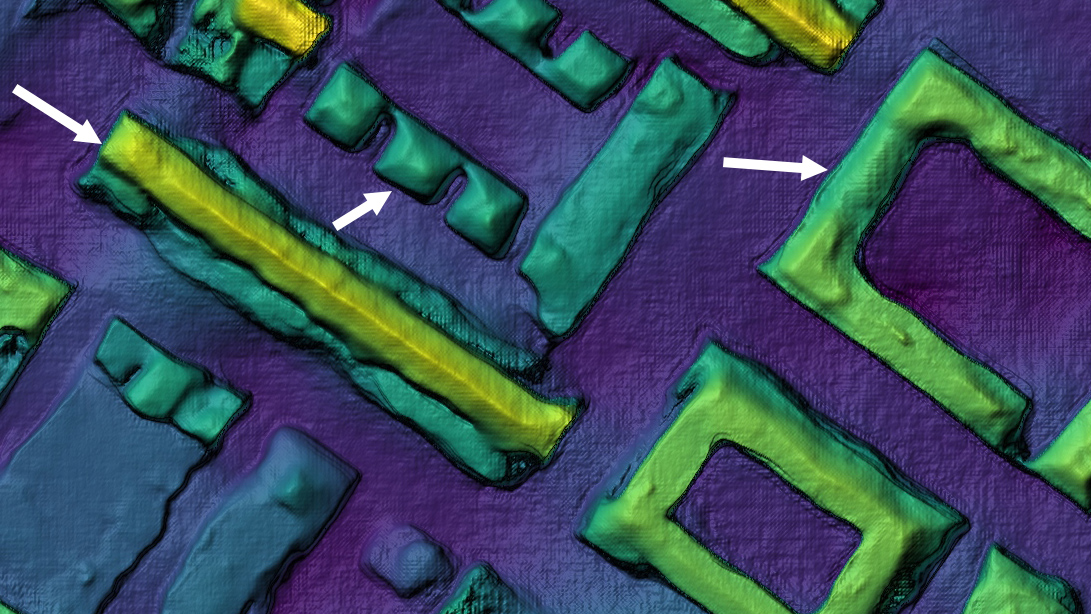}
        \\
        \rotatebox[origin=c]{90}{\footnotesize \ours-mono} &
        \rotatebox[origin=c]{90}{\footnotesize (ours)} &
        \includegraphics[width=\mywidth,trim={0 0 0 0},clip]{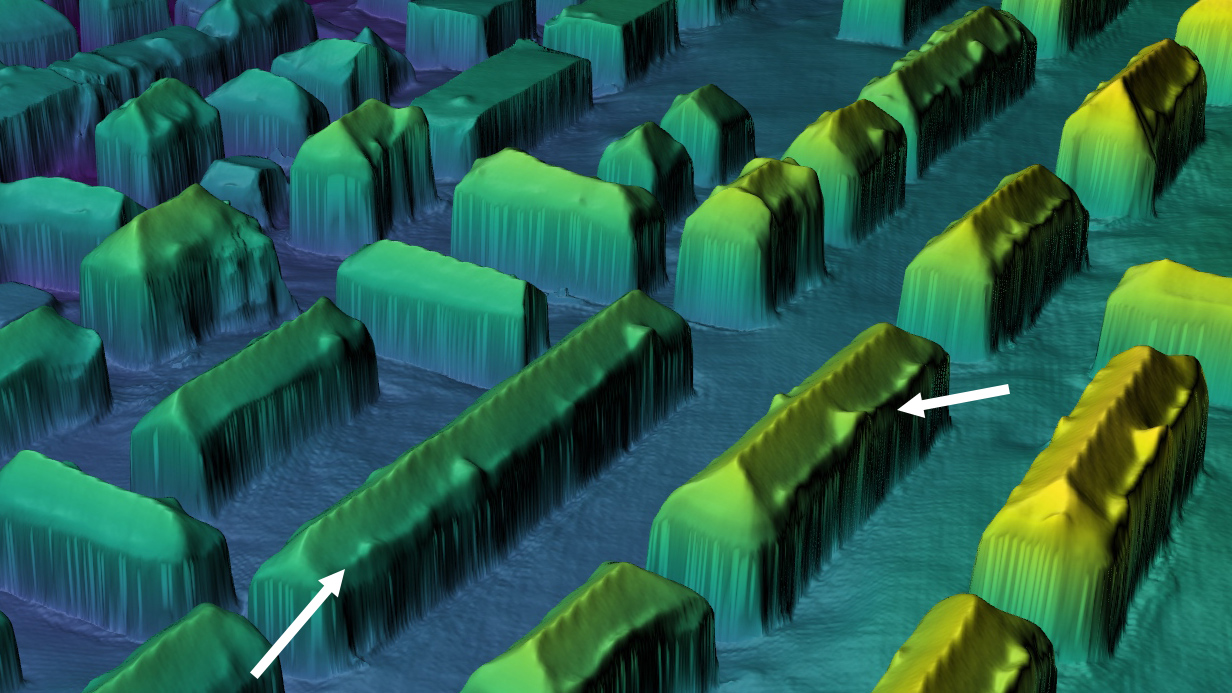} &
        \includegraphics[width=\mywidth,trim={0 0 0 0},clip]{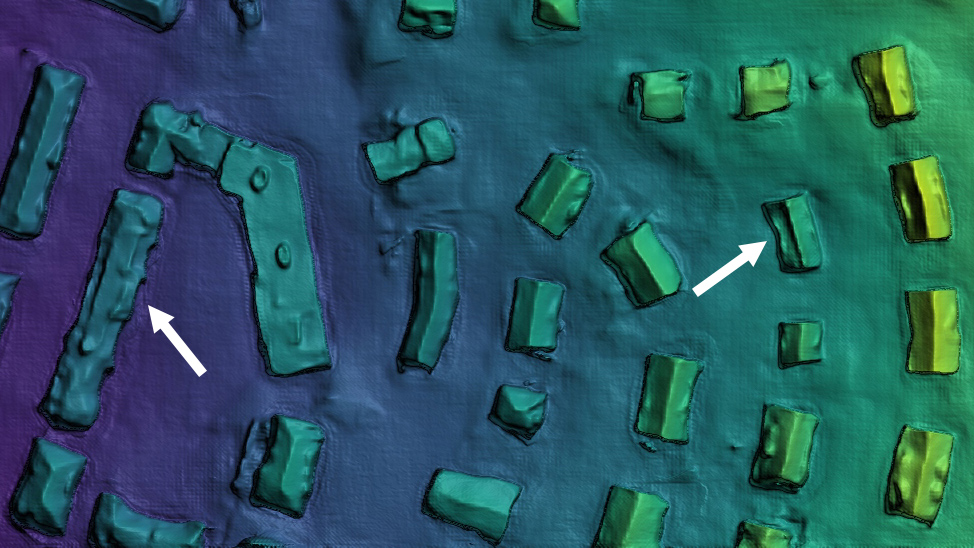} &
        \includegraphics[width=\mywidth,trim={0 0 0 0},clip]{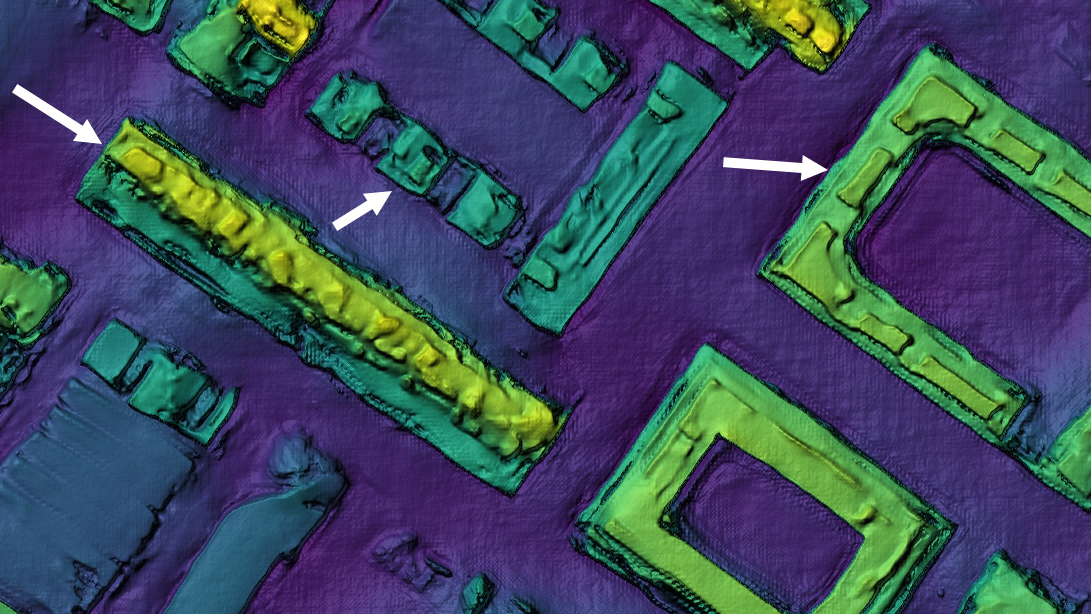}
        \\
        \rotatebox[origin=c]{90}{\footnotesize \ours-stereo} &
        \rotatebox[origin=c]{90}{\footnotesize (ours)} &
        \includegraphics[width=\mywidth,trim={0 0 0 0},clip]{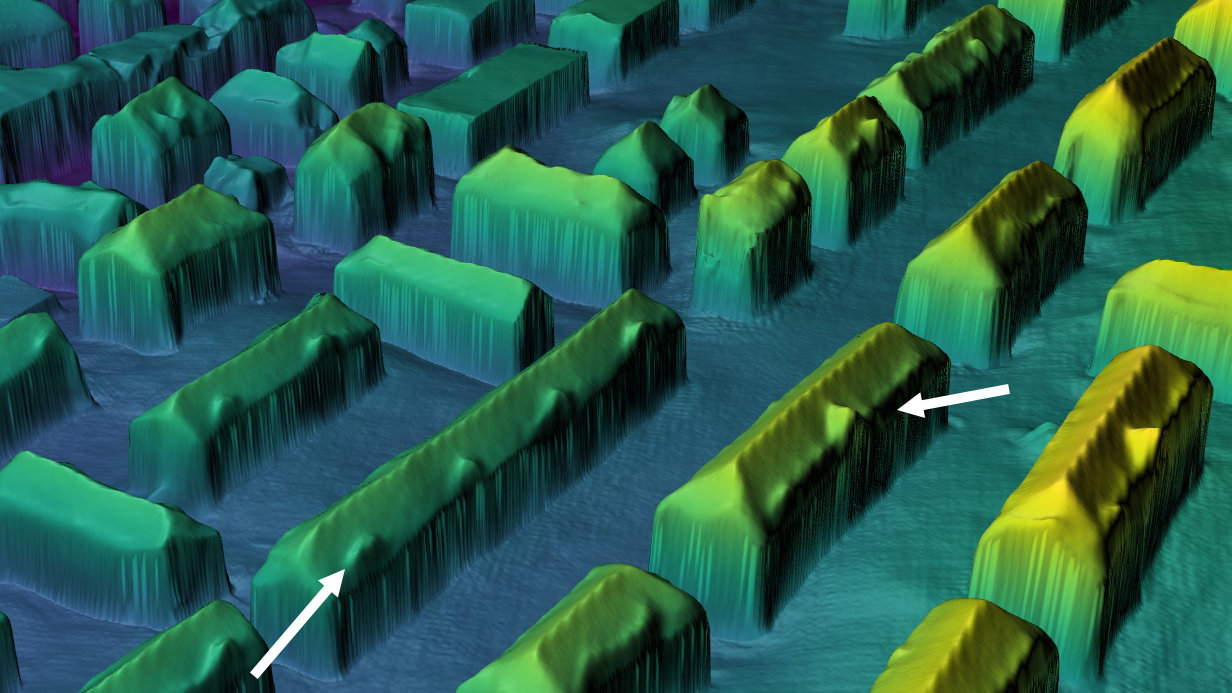} &
        \includegraphics[width=\mywidth,trim={0 0 0 0},clip]{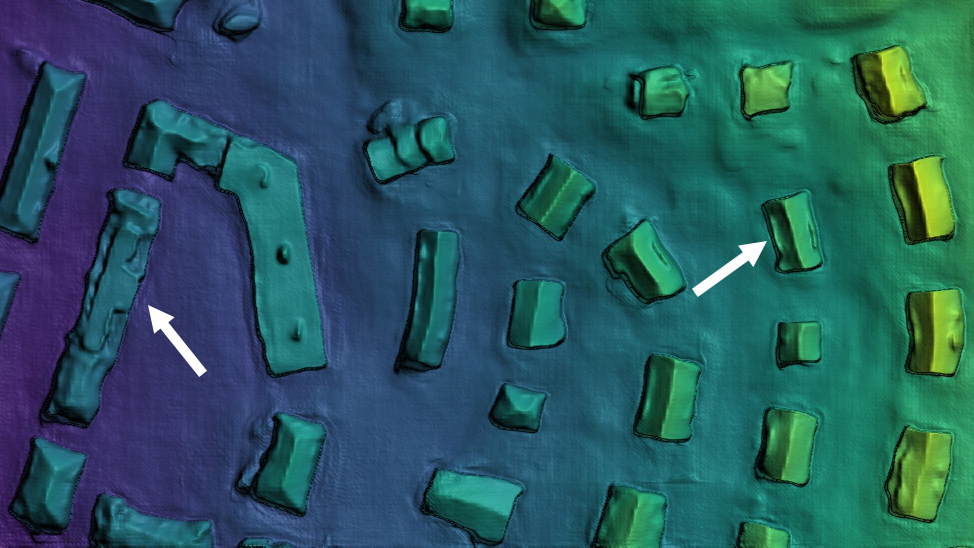} &
        \includegraphics[width=\mywidth,trim={0 0 0 0},clip]{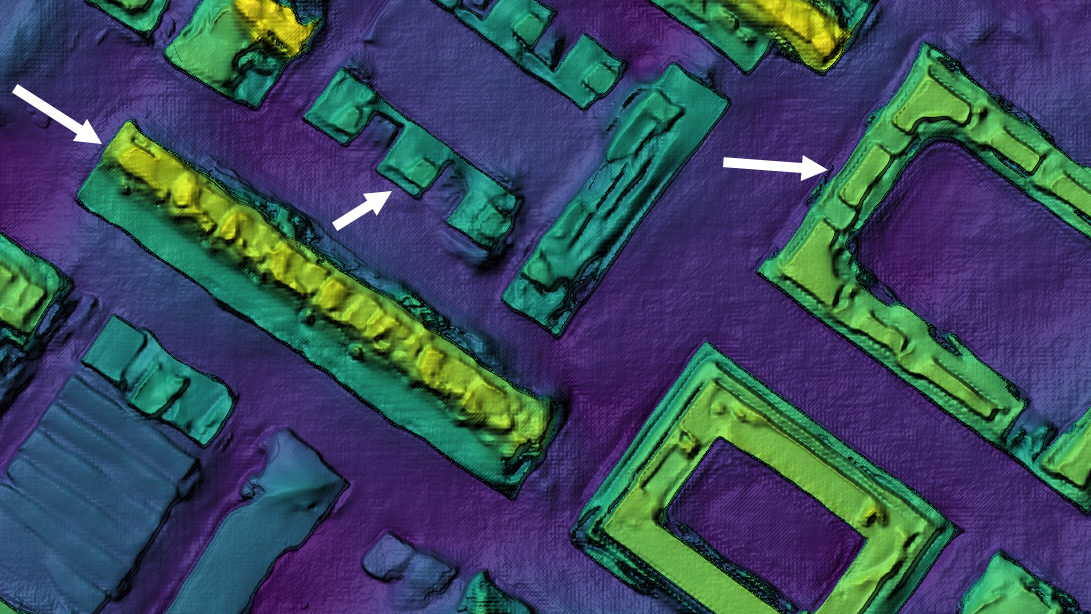}
        \\
        \rotatebox[origin=c]{90}{\footnotesize Google Earth view} &  &
        \includegraphics[width=\mywidth,trim={0 0 0 0},clip]{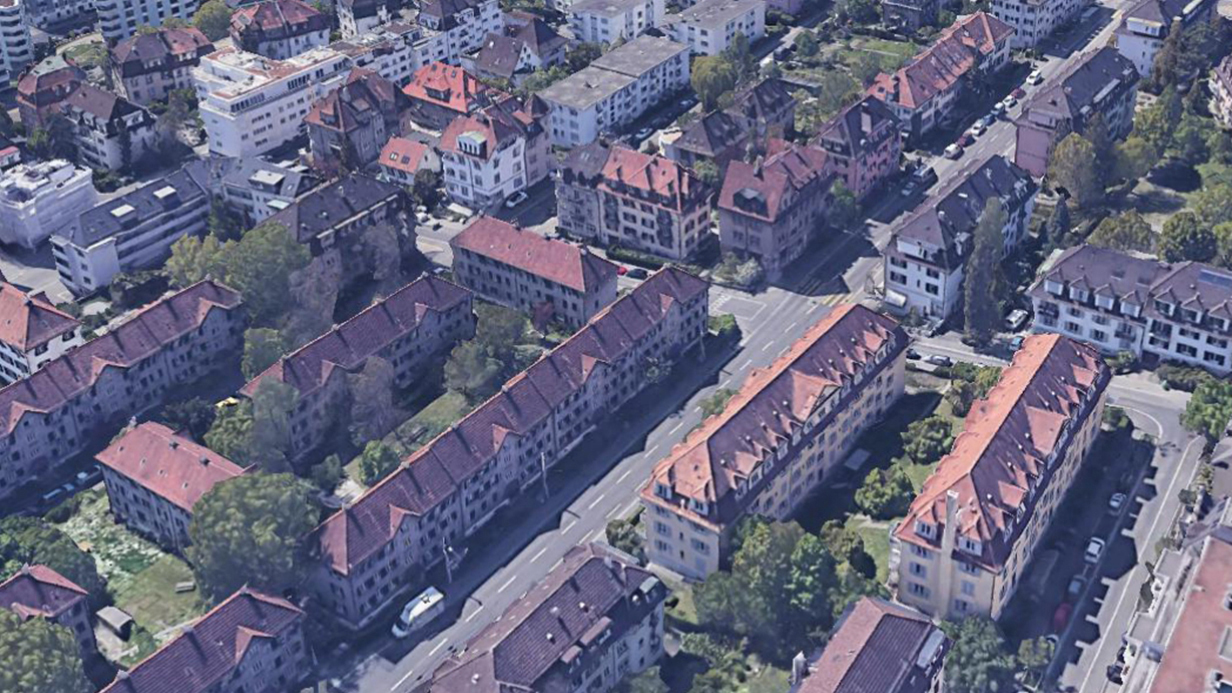} & 
        \includegraphics[width=\mywidth,trim={0 0 0 0},clip]{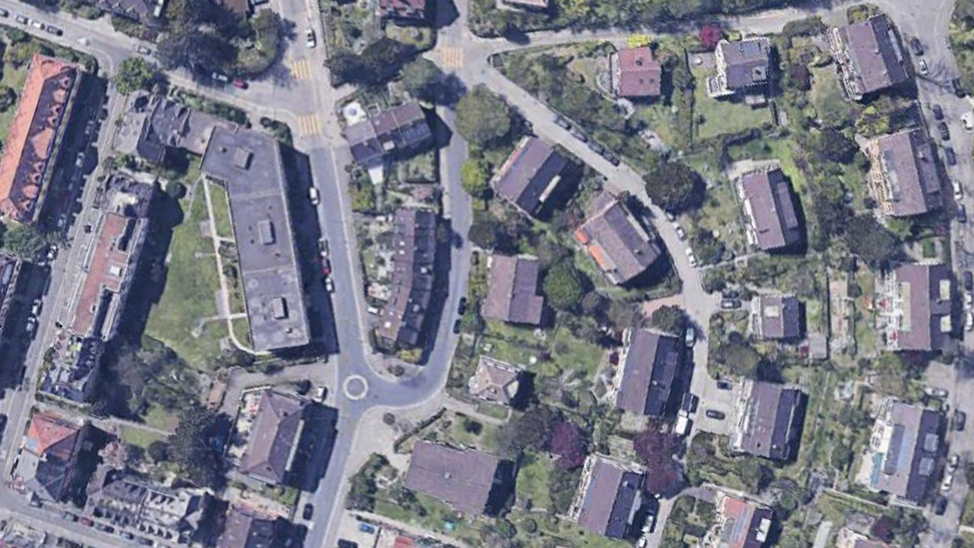} & 
        \includegraphics[width=\mywidth,trim={0 0 0 0},clip]{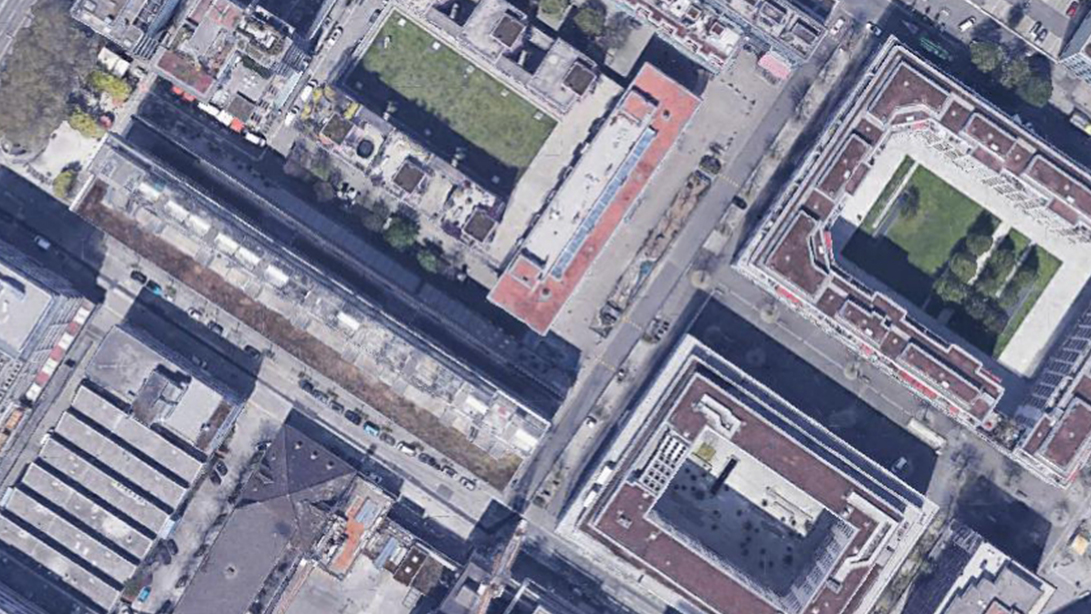}
        \\
    \end{tabular}
    \vspace{-0.5em}
    \caption{Visual comparison of different \ours variants with selected baselines. Heights are color-coded from blue to green to yellow. All examples are from the test set.}
    \label{fig:qualitative_results}
\end{figure*}

In the following, we analyze the performance of \ours and compare it to the baselines described above. We present quantitative results in Table~\ref{tab:ablation} and visual examples in Figure~\ref{fig:qualitative_results}.

\mypara{\ours-0}
We start by assessing the performance of \mbox{\ours}-0, so as to quantify the impact of implicit neural modeling in isolation, without image guidance. The DSM generated by conventional photogrammetric processing serves as the baseline. Recall, that DSM was generated by rasterizing the point cloud ~$\mathcal{P}$  with a cell-wise median of the highest 3D points, followed by standard denoising (cf.\ Section~\ref{sec:data}). Due to the limited resolution and inhomogeneous radiometry of satellite imagery, the resulting DSM is fairly noisy and lacks sharp features (Fig.~\ref{fig:qualitative_results}, \nth{1}~row). The MAE is $\approx\,$3.9$\,$m and the MedAE 1.6$\,$m. The RMSE is almost 2$\times$ higher than the MAE because of a small number of substantial matching errors. Applying \ours-0 to the point cloud $\mathcal{P}$ improves the reconstruction significantly. The MAE is lowered to $\approx\,$1.9$\,$m, an improvement of \textgreater50\% compared to the conventional, heuristic procedure. Similarly, the RMSE and MedAE are also reduced to 3.6$\,$m, respectively 0.9$\,$m. We note that vegetation is not included in the ground truth. Therefore, \ours-0 learns to filter out vegetation, which naturally leads to improved quality metrics. Nonetheless, even when excluding forested areas from the evaluation, we observe a decrease in MAE of almost 43\% to 1.6$\,$m for general terrain. For buildings, the MAE drops by 25\% to 2.3$\,$m. Visually, the DSM generated with \ours-0 exhibits markedly less surface noise, and built-up areas are separated into plausible individual units (Fig.~\ref{fig:qualitative_results}, \nth{5}~row). Despite the limited quality and sparsity of the input point cloud, \mbox{\ours-0} reconstructs sharp building edges and gable roofs and even recovers buildings that are hardly visible in the conventional DSM (Fig.~\ref{fig:qualitative_results}, \nth{5}~row, \nth{2}~column). Building footprints are, however, somewhat wobbly, and visually discernible roof details are missed entirely (Fig.~\ref{fig:qualitative_results}, \nth{5}~row, \nth{3}~column).

\mypara{Influence of Image Guidance}
In addition to the shape code $\psi(\mathcal{P}, \mathbf{x})$, \ours-mono and \ours-stereo exploit a pixel-aligned image code $\xi(\mathcal{I}, \mathbf{x})$ to guide the occupancy prediction. In this way, both image-guided network variants can, on the one hand, potentially capture 
\pagebreak
general correlations between image patterns and the underlying surface shape and, on the other hand, learn to precisely align the reprojected 3D scene geometry with 2D image discontinuities. \mbox{\ours-mono}, leveraging a single ortho-image to generate~$\xi$, boosts the reconstruction accuracy by a significant margin compared to \mbox{\ours-0}. The overall MAE is decreased by 0.3$\,$m to $\approx\,$1.6$\,$m, and the MedAE by 0.2$\,$m to 0.7$\,$m. With a relative improvement of 18\%, we observe the largest gain in accuracy for terrain. For buildings, all metrics improve by $\approx\,$12\%. Qualitatively, \ours-mono produces sharper and straighter building outlines and roof features, reconstructs buildings missed by \ours-0, and partially recovers geometric details on roofs (Fig.~\ref{fig:qualitative_results}, \nth{6}~row, \nth{1}~column). \mbox{\ours}-stereo, using a second image to generate the latent code~$\xi$, further improves the reconstruction. While the quantitative gains are rather small, the visual quality of the reconstructed 3D geometry is clearly enhanced. Buildings have crisp roof lines, and there are fewer implausible bumps on the terrain. Perhaps most striking is the recovery of fine-grained roof structures like dormers (Fig.~\ref{fig:qualitative_results}, \nth{7}~row, \nth{1}~column). Our full model achieves a MAE of 1.9$\,$m for buildings and 1.3$\,$m for terrain.

\mypara{Comparison to Learning-based Baselines}
Among all methods that do not have access to monocular or stereo information, \ours-0 yields the lowest reconstruction errors (Table~\ref{tab:ablation}, rows~\mbox{2--4}). Compared to \resdepth-0, it reconstructs smoother surfaces and more accurate building heights and roof features (Fig.~\ref{fig:qualitative_results}, \nth{2} and \nth{5}~row). Quantitatively, the difference amounts to 15\% in overall MAE and to 20\% in MAE for buildings. On the terrain, all methods perform comparably, except that \resdepth-0 is a bit worse in forest regions. We speculate that the differences are mostly due to the network input. \mbox{\ours-0} receives a rich shape code~$\psi$ that encodes the local distribution of the input point cloud and potentially includes some information about the vertical point distribution, whereas \resdepth-0 has no access to such a \say{vertical height profile} and must infer the height entirely from 2.5D local context.
PIFu-0, a generic neural model for implicit 2.5D height field reconstruction, performs slightly better than \mbox{\resdepth-0}. It achieves an overall MAE of 2.0$\,$m, which is 10\% lower than \resdepth-0. The gain is primarily caused by (overly) smoothed surfaces and more accurate terrain estimates in vegetated areas. Compared to \ours-0, the overall MAE is 6\% higher and the building MAE is 22\% higher.

With additional (monocular or stereo) image information to support the reconstruction, we observe the same trend across all methods. The variants using a single image (Table~\ref{tab:ablation}, rows~\mbox{5--7}) outperform their respective counterparts 
\pagebreak
without image guidance by \mbox{16--25\%} in terms of overall accuracy. Stereo-enabled variants (Table~\ref{tab:ablation}, rows~\mbox{8--10}) bring another improvement of \mbox{4--7\%} compared to the monocular versions. Notably, we find the biggest relative gain for \resdepth and the smallest one for \ours. We hypothesize that, despite having co-registered latent embeddings, it may be more difficult to discover correlations between separate image codes $\xi(\mathcal{I}, \mathbf{x})$  and shape codes $\psi(\mathcal{P}, \mathbf{x})$ than to find correlations between ortho-photos and the height raster when directly stacked into a multi-channel image. In terms of absolute metrics, \ours-0 and \ours-mono are superior to the respective \resdepth and PIFu variants. When using stereo images as guidance, the quantitative performance of all three methods is very similar. Nevertheless, in terms of visual quality, \ours-stereo is the sole method capable of recovering small roof details like dormers on single-family houses, see Figure~\ref{fig:qualitative_results}.

\mypara{Computational Complexity}
Implicit shape models like \mbox{\ours} have comparatively high computational cost at inference time. For every DSM patch, one must \textit{(i)}~perform one forward pass through the shape and image encoders to build the latent embeddings $\psi$ and $\xi$; and \textit{(ii)}~for every query point~$\mathbf{x}$ run a forward pass through the decoder to retrieve the occupancy. With our current implementation, which has not yet been tuned for speed, the occupancy decoding takes roughly 3$\times$ longer than the feature encoding, and the entire reconstruction needs \mbox{$\approx\,$9~mins$\,$/$\,$km\textsuperscript{2}}, not counting the preceding ortho-rectification of the images. In comparison, inference with \resdepth only requires a single U-Net forward pass per DSM patch, with a run time \mbox{$<\,$5~sec$\,$/$\,$km\textsuperscript{2}}. To put the computation times in perspective, note that even our unoptimized implementation of \ours will take only $\approx\,$15 hours to process a city of \mbox{100$\,$km\textsuperscript{2}} on a single machine and would not constitute a major bottleneck of the overall reconstruction pipeline.

\section{Conclusion}
We have presented \ours, a method that creates DSMs from raw photogrammetric point clouds and ortho-images with the help of an implicit neural 3D scene representation. \mbox{\ours} is able to reconstruct DSMs at city scale and, in our experiments, reduces the MAE by \textgreater60\% compared to conventional stereo pipelines. In comparison with learned DSM refinement schemes, \ours is particularly good at recovering minute shape details such as dormers and produces exceptionally crisp and straight building edges.
Interesting future research directions include how to encode point clouds and multi-view images jointly in a single latent representation rather than separately; and extracting 
\pagebreak
full 3D surfaces from the implicit representation rather than 2.5D DSMs. \mbox{\ours} has so far been validated under ideal machine learning conditions, with training and test regions that lie next to each other and have been observed in the same satellite images. Further work is needed to assess its ability to generalize across variations in stereo geometry, image radiometry, and geographic context. In the extended technical report \cite{stucker2022implicity}, we provide first, exploratory experiments for geographical generalization.

{
	\begin{spacing}{1.17}
		\normalsize
		\bibliography{bib}
	\end{spacing}
}

\renewcommand{\thefigure}{A\arabic{figure}}
\setcounter{figure}{0}
\renewcommand{\thetable}{A\arabic{table}}
\setcounter{table}{0}
\setcounter{section}{0}

\def\thesection{\Alph{section}}

\twocolumn[
    \begin{center}
        \vspace{-0.7em}
        {\normalsize\bf\expandafter\uppercase\expandafter{\large\ours: City Modeling from Satellite Images with\\Deep Implicit Occupancy Fields} \par}%
        \vskip 1em
        {\normalsize\bf\expandafter{Supplementary Material} \par}%
        \vspace*{18pt}
        {\normalsize
        \begin{tabular}[t]{c}
            Corinne Stucker, Bingxin Ke, Yuanwen Yue, Shengyu Huang, Iro Armeni, Konrad Schindler \vspace{1mm}\\
            ETH Zurich, Switzerland
        \end{tabular}
        }
        \vskip 2\baselineskip
        \vskip 3ex
    \end{center}
]

\section{Geographical Generalization\\Within Zurich}
So far, \ours has been validated under optimal machine learning conditions, where training and test regions share the same urban style, topography, and imaging conditions (viewing directions, stereo geometry, lighting, atmospheric conditions). Such a setting with hardly any domain shift may already be of use in certain scenarios,\footnote{E.g., to periodically reconstruct a fixed geographic region monitored from the same orbits.} but the learned prior will be tailored to the specific characteristics of the training data. Consequently, one would expect the performance to deteriorate with increasing domain gap between training and test sites, due to variations in viewing directions, stereo geometry, and architectural layout. To train a learned system that can operate at large scale, additional measures are required that enhance invariance against those sources of variability, as is demonstrated, for instance, in \cite{stucker2022resdepth}.

As a first step towards quantifying the generalization performance of \ours, we evaluate its ability to generalize across space, without altering the image set used to generate the network inputs. This scenario is representative to assess whether the model learns generic features that remain valid for DSM reconstruction in architecturally not too different, but previously unseen locations; provided that the training and test data (point clouds and ortho-photos) share the same characteristics.

\begin{figure}[!ht]
    \centering
    \includegraphics[trim={0 0 0 0}, clip,width=\columnwidth]{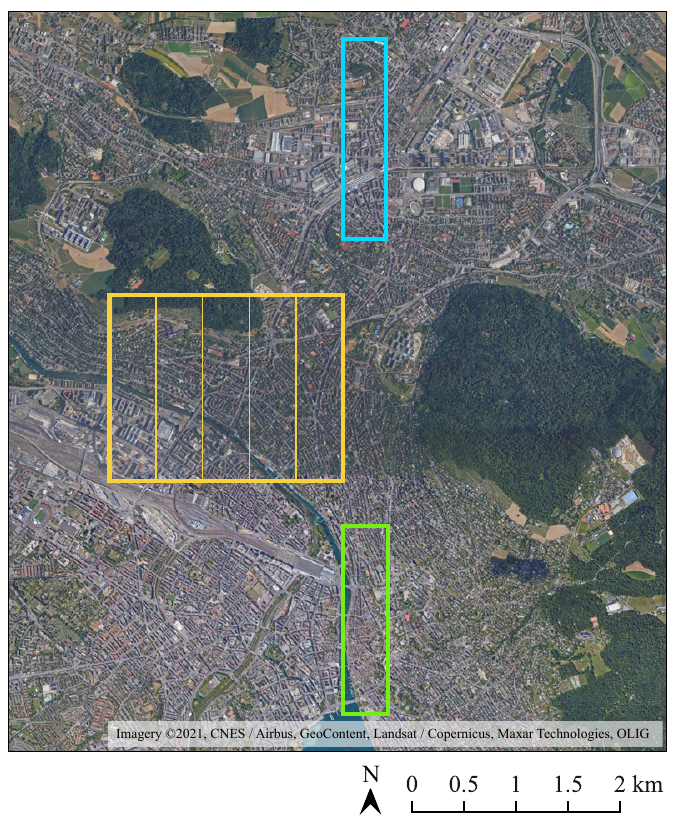}
    \caption{Study areas in Zurich. We define three areas \zurichOne (yellow), \zurichTwo (green), and \zurichThree (blue), where \zurichOne is split into separate stripes for training, validation, and testing. We use \zurichOne to perform the experiments in the main paper and \zurichTwo and \zurichThree to test geographical generalization across the city.}
    \label{fig:ROI_ZUR}
\end{figure}

\mypara{Study areas}
We select three geographic areas to quantify the generalization performance of \ours across different districts of Zurich. The areas are referred to as \zurichOne, \zurichTwo, and \zurichThree and are shown in Figure~\ref{fig:ROI_ZUR}. \zurichOne corresponds to the study area used for the experiments in the main paper and is divided into five equally large, mutually exclusive stripes for training, validation, and testing. It covers 4{$\,$km\textsuperscript{2}} and contains widely spaced, detached residential buildings, allotments, and high commercial buildings. Furthermore, it includes a forested hill and stretch of the river Limmat. With an extent of 0.8{$\,$km\textsuperscript{2}}, \zurichTwo and \zurichThree are both equally large as the test stripe in \zurichOne. \zurichTwo is located in the heart of Zurich and primarily covers the historic city center. \zurichThree comprises low residential districts, moderately high buildings, and some hilly, forested terrain.

\mypara{Experimental Setup}
We train \ours-stereo using the three training stripes of \zurichOne, following the same training procedure and settings as in the main paper. We then employ the learned model to reconstruct DSMs of \zurichTwo and \zurichThree, without fine-tuning or any other measures to support geographic generalization. For inference, we resort to the same pair of ortho-images also used during training, i.e., the behavior of \ours for a more distant location is separated from the impact of varying imaging conditions.

\mypara{Baselines}
We compare the performance of \ours-stereo against \resdepth-stereo, a learned DSM refinement approach based on an explicit 2D scene representation, and also using stereo guidance~\cite{stucker2022resdepth}. In analogy to \mbox{\ours}, we train \resdepth on the three training stripes of \zurichOne and apply the trained model to \zurichTwo and \zurichThree, without further fine-tuning. For a fair comparison, we use the exact same ortho-images as for \mbox{\ours}. As a second baseline, we also compare against the conventional stereo DSM, without any learned refinement (see Section~\ref{sec:data} of the main paper for details).

\mypara{Results}
\begin{table*}[!htp]
    \setlength\dashlinedash{2pt}
    \setlength\dashlinegap{1.5pt}
    \setlength\arrayrulewidth{0.3pt}
    \centering
    \begin{adjustbox}{max width=0.95\textwidth}
        \begin{tabular}{@{\hspace{\tabcolsep}}llccccccccc@{\hspace{\tabcolsep}}}
			\toprule
			Test area & Reconstruction & \multicolumn{3}{c}{Overall} & \multicolumn{3}{c}{Buildings} & \multicolumn{3}{c}{Terrain} \\
			\cmidrule(lr){3-5}\cmidrule(lr){6-8}\cmidrule(lr){9-11}
			& & MAE & RMSE & MedAE & MAE & RMSE & MedAE & MAE & RMSE & MedAE \\
			\midrule
			
			\multirow{3}{*}{\zurichOne} & Conventional DSM  & 	 3.89 & 7.03 & 1.59 &    3.02 & 5.02 & 1.47 &    4.29 & 7.78 & 1.65 \\
			& \resdepth-stereo    &    1.53 &  2.97 &  0.74 &     1.91 &  3.93 &  0.82 &    1.35 &  2.41 &  0.71 \\
			& \ours-stereo &      1.52 &  2.91 &  0.70 &     1.93 & 3.86 &  0.78 &    1.33 &  2.35 &  0.67 \\
			\hdashline\noalign{\vskip 0.7ex}
			\multirow{3}{*}{\zurichTwo} & Conventional DSM &    3.92 &  5.73 &  2.24 &     3.34 &  5.05 &  1.98 &    4.36 &  6.18 &  2.51 \\
			& \resdepth-stereo &    2.48 &  3.81 &  1.69 &     2.91 &  4.65 &  1.80 &    2.15 &  3.03 &  1.61 \\
			& \ours-stereo &    2.03 &  3.53 &  1.00 &     2.32 &  4.22 &  0.95 &    1.81 &  2.91 &  1.05 \\
            \hdashline\noalign{\vskip 0.7ex}
			\multirow{3}{*}{\zurichThree} & Conventional DSM  &    3.65 &  5.87 &  2.02 &     4.63 &  6.76 &  2.72 &    3.22 &  5.44 &  1.82 \\
			& \resdepth-stereo &    2.90 &  4.26 &  2.25 &     4.28 &  6.48 &  2.60 &    2.29 &  2.75 &  2.14 \\
			& \ours-stereo &    2.55 &  3.82 &  1.96 &     3.61 &  5.69 &  2.22 &    2.08 &  2.60 &  1.87 \\
		    \bottomrule
	    \end{tabular}
    \end{adjustbox}
    \caption{Geographical generalization across space within Zurich. We report the mean absolute error (MAE), the root mean square error (RMSE), and the median absolute error (MedAE) over all pixels, building pixels, and terrain pixels. All values are meters.\\The evaluation of the test area in \zurichOne is identical to the results in the main paper.}
    \label{tab:generalization_ZUR}
\end{table*}

\begin{figure*}[!htpb]
    \centering
    \subfigure[Conventional reconstruction]{
        \begin{minipage}[t]{0.8\textwidth}
            \centering
            \includegraphics[width=0.45\textwidth]{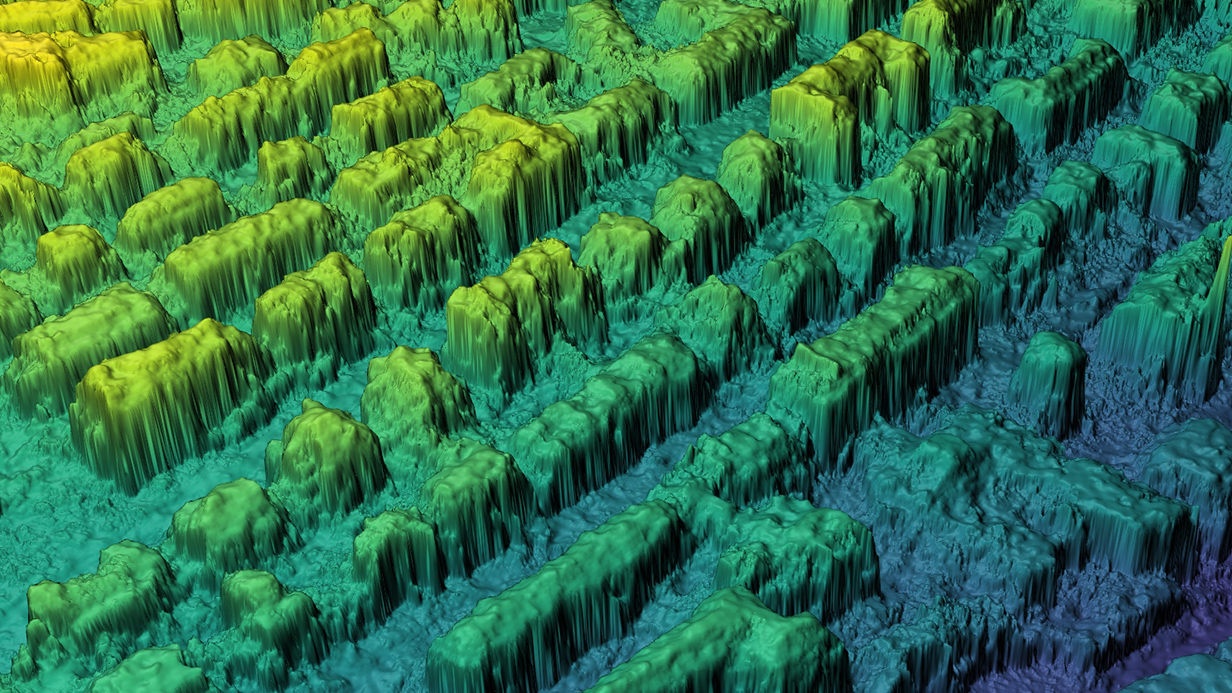}
            \hspace{0.5em}
            \includegraphics[width=0.45\textwidth]{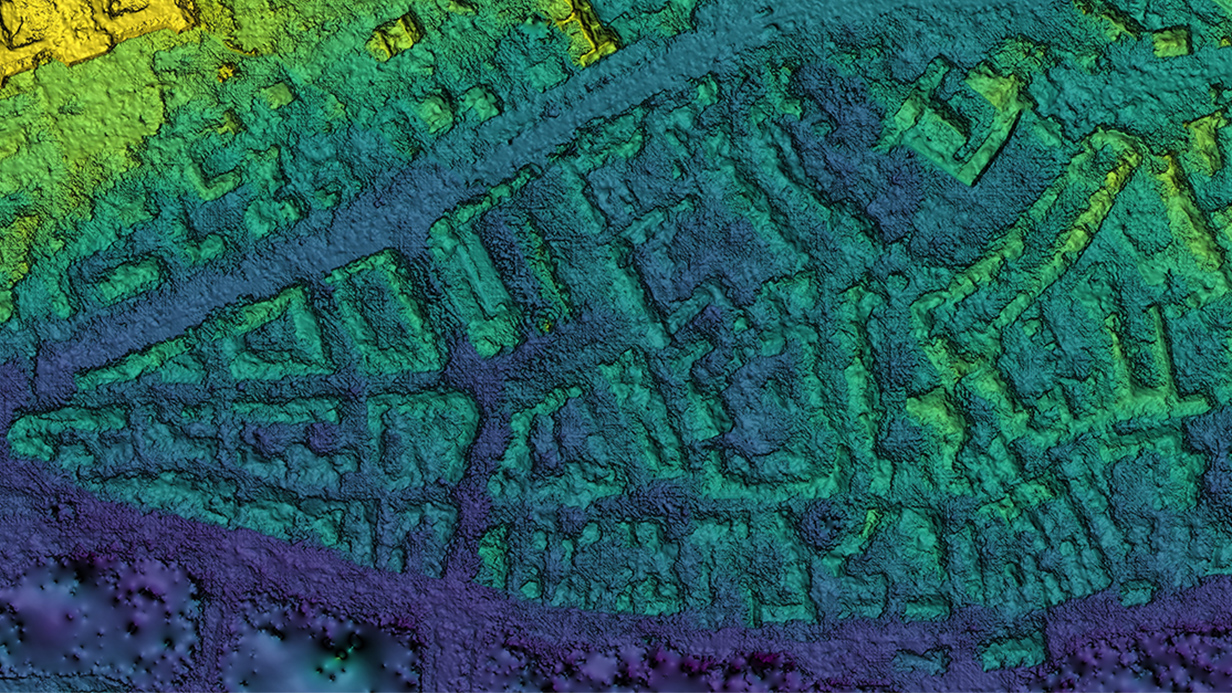}\vspace{2mm}
        \end{minipage}
    }
    \\
    \subfigure[\resdepth-stereo]{
        \begin{minipage}[t]{0.8\textwidth}
            \centering
            \includegraphics[width=0.45\textwidth]{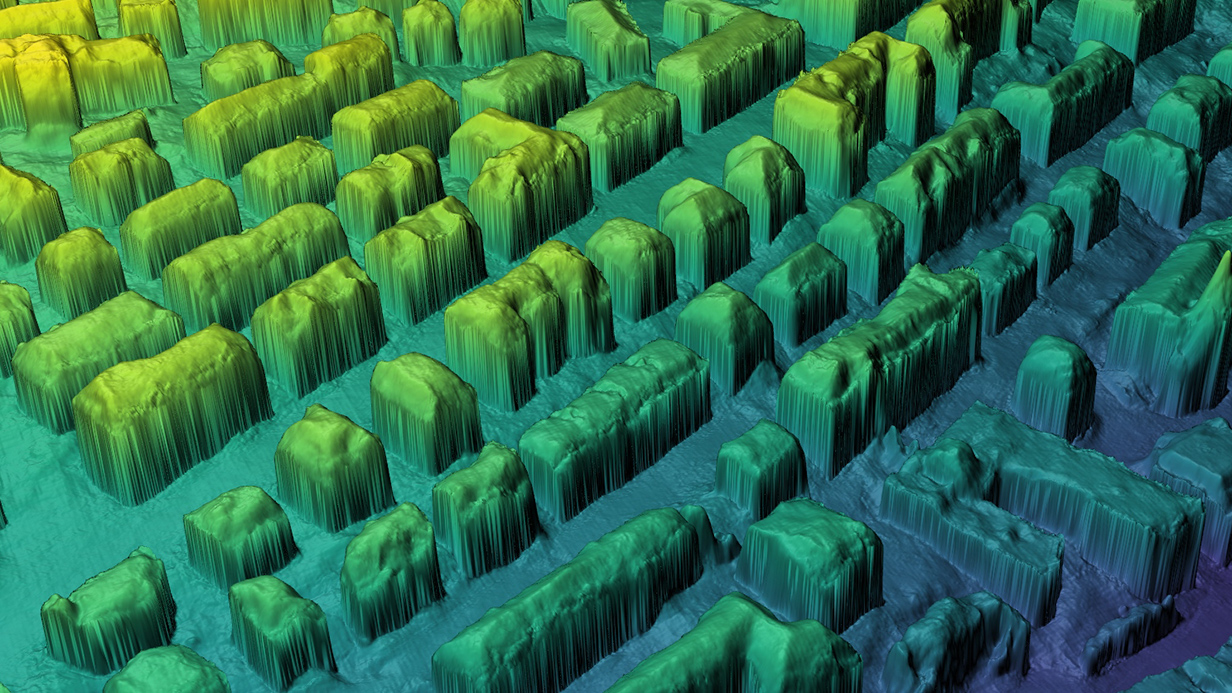}
            \hspace{0.5em}
            \includegraphics[width=0.45\textwidth]{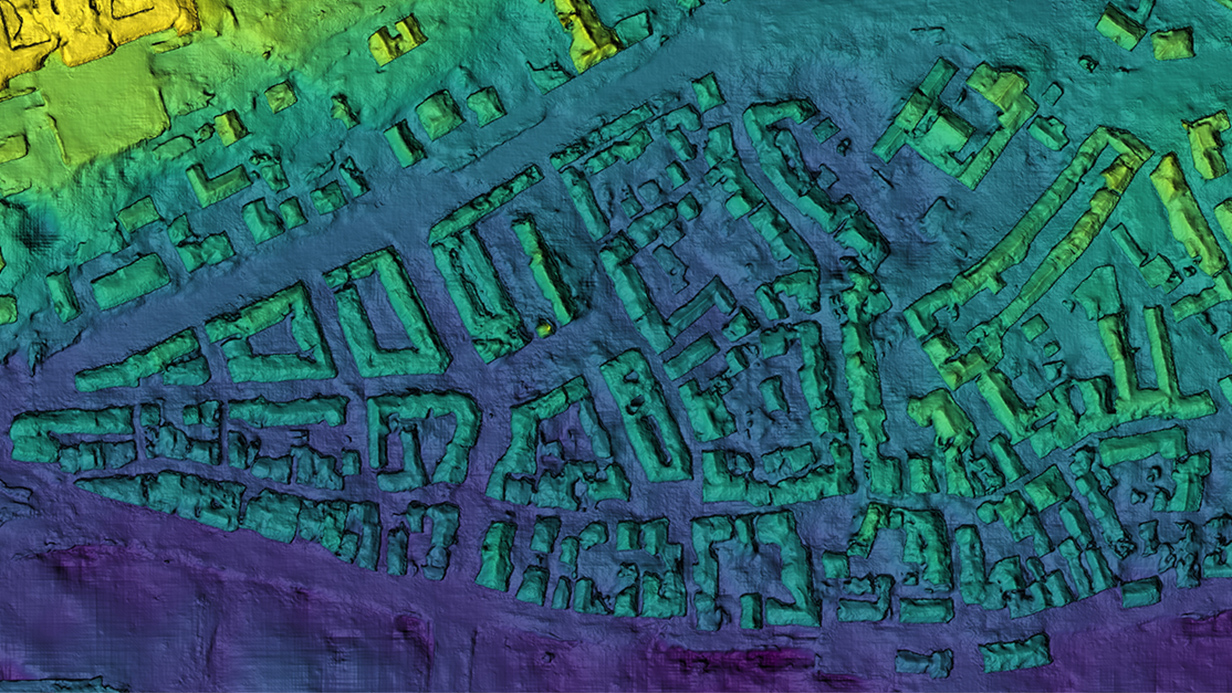}\vspace{2mm}
        \end{minipage}
    }
    \\
    \subfigure[\ours-stereo (ours)]{
        \begin{minipage}[t]{0.8\textwidth}
            \centering
            \includegraphics[width=0.45\textwidth]{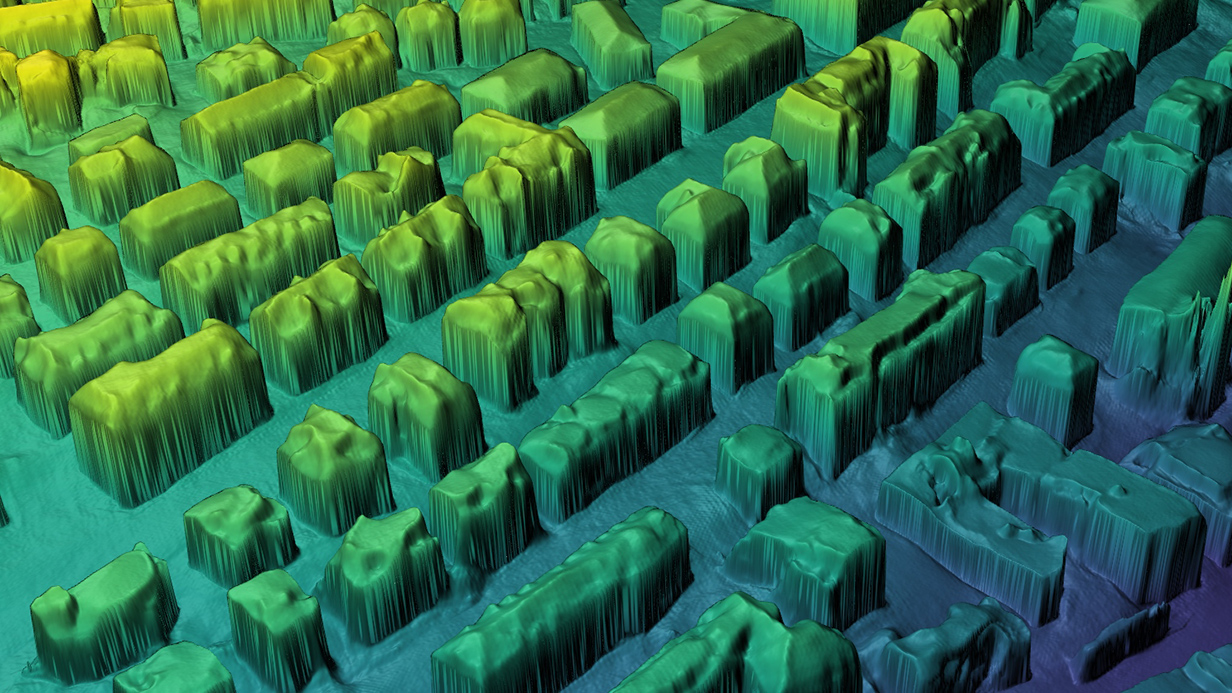}
            \hspace{0.5em}
            \includegraphics[width=0.45\textwidth]{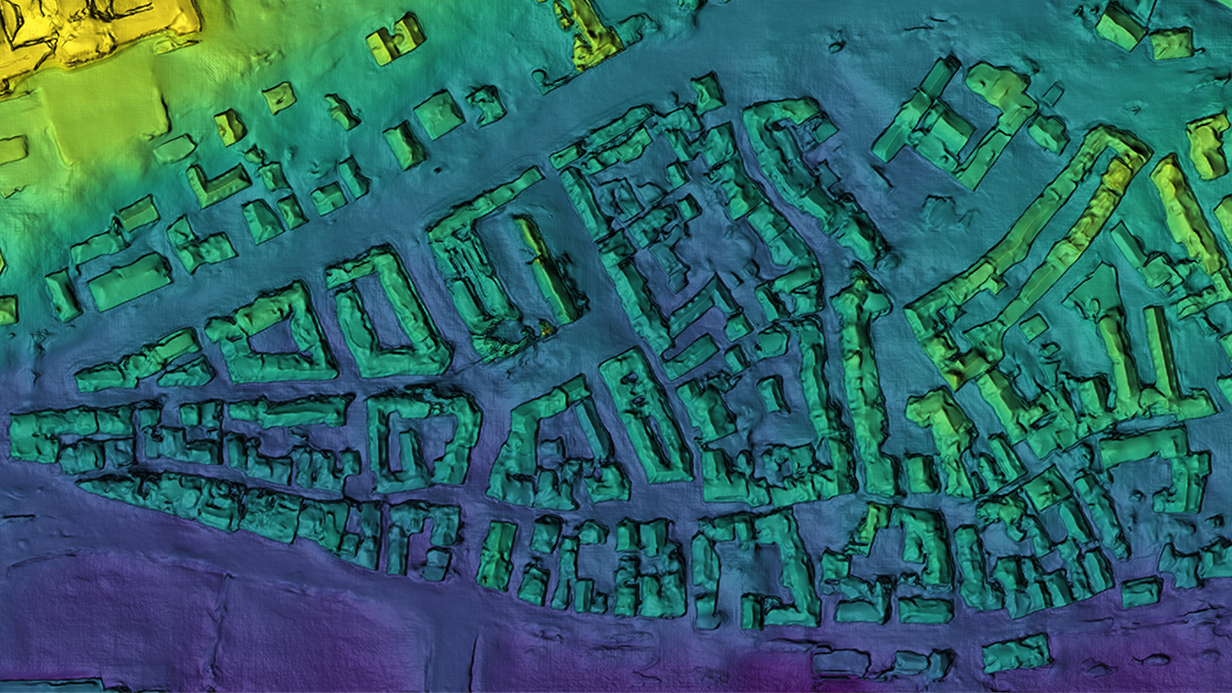}\vspace{2mm}
        \end{minipage}
    }
    \\
    \subfigure[Google Earth view (left) and example satellite view (right)]{
        \begin{minipage}[t]{0.8\textwidth}
            \centering
            \includegraphics[width=0.45\textwidth]{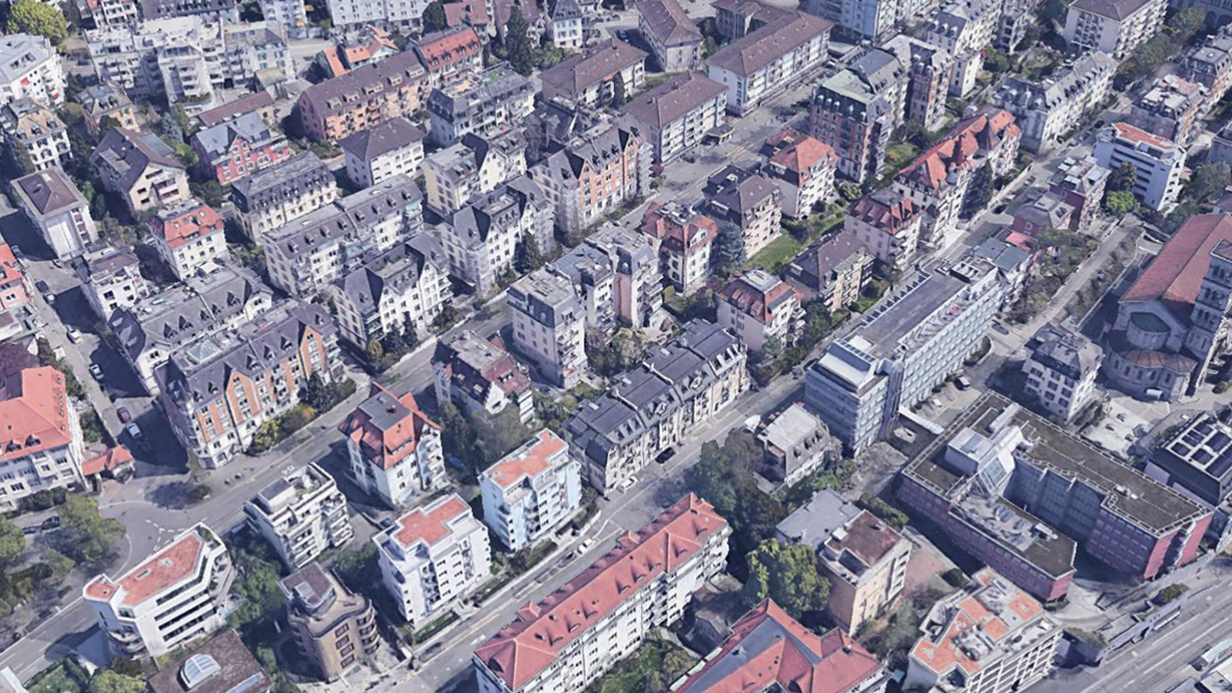}
            \hspace{0.5em}
            \includegraphics[width=0.45\textwidth]{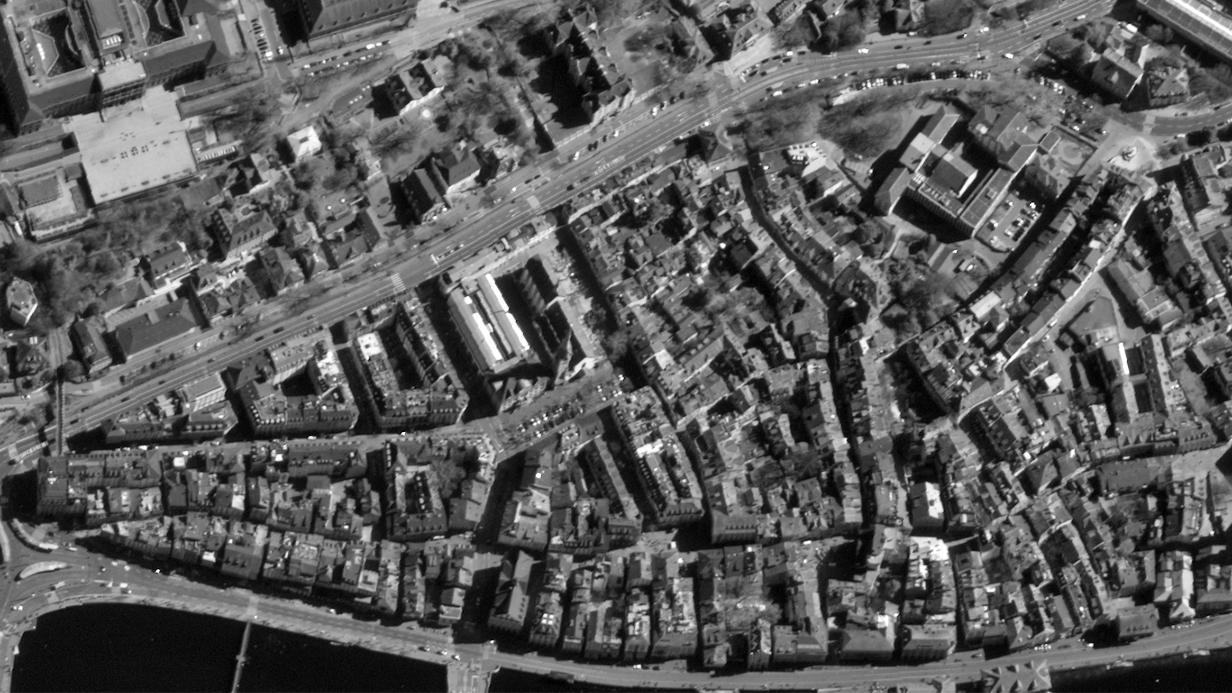}\vspace{2mm}
        \end{minipage}
    }
    \vspace{-0.5em}
    \caption{Geographical generalization between different regions of Zurich. Heights are color-coded from blue to green to yellow.\\We train \resdepth-stereo and \ours-stereo in \zurichOne (see Figure~\ref{fig:ROI_ZUR}) and apply the trained models without fine-tuning in \zurichTwo. Reconstruction results in the north of of \zurichTwo (left) and for the historic city center (right).}
    \label{fig:generalization_ZUR2}
\end{figure*}
\begin{figure*}[!ht]
    \centering
    \subfigure[Conventional reconstruction]{\includegraphics[width=0.23\textwidth]{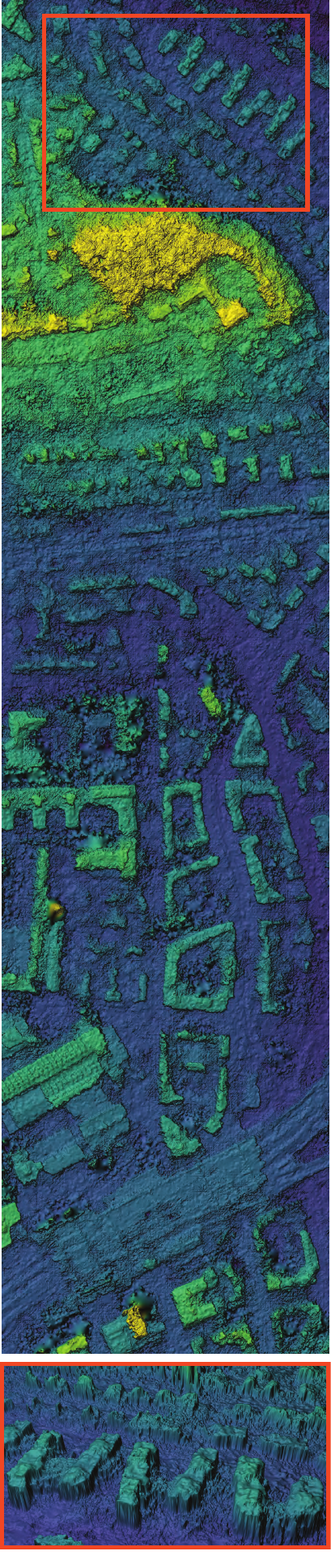}}
    \hfil
    \subfigure[\resdepth-stereo]{\includegraphics[width=0.23\textwidth]{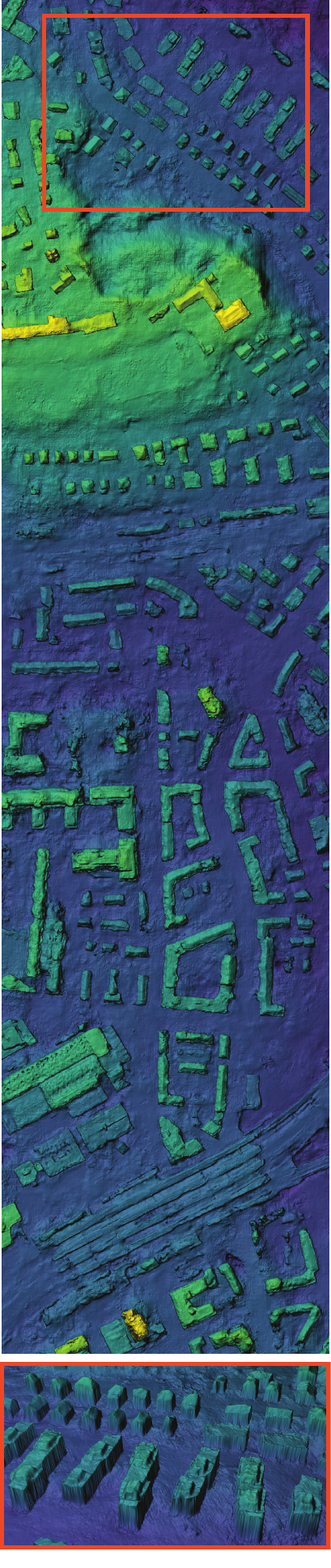}}
    \hfil
    \subfigure[\ours-stereo (ours)]{\includegraphics[width=0.23\textwidth]{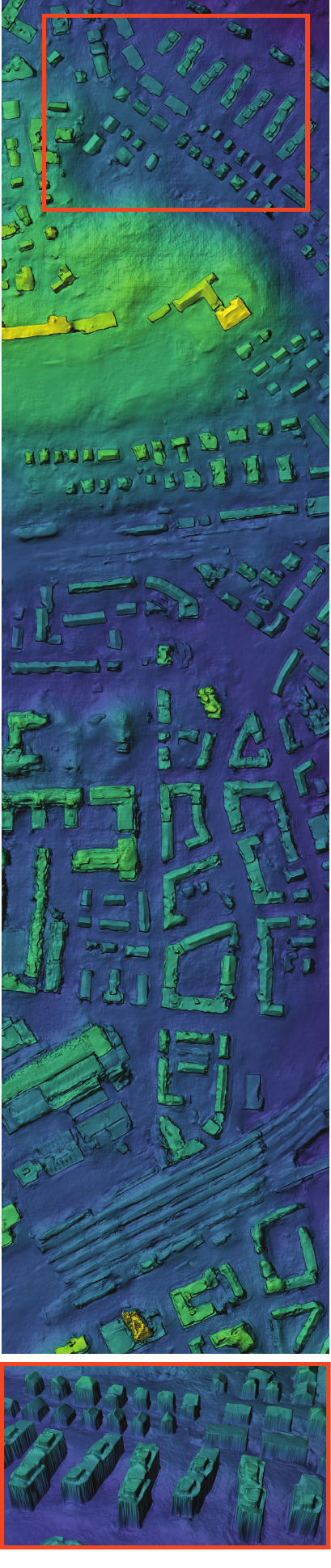}}
    \hfil
    \subfigure[Example satellite view (top) and Google Earth view (bottom)]{\includegraphics[width=0.23\textwidth]{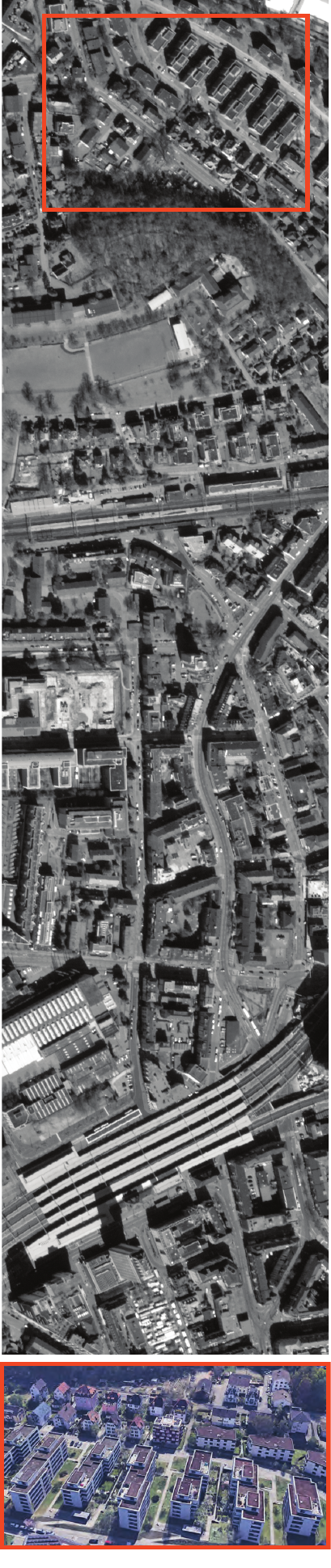}}
    \vspace{-0.5em}
    \caption{Geographical generalization between different regions of Zurich. Heights are color-coded from blue to green to yellow.\\We train \resdepth-stereo and \ours-stereo in \zurichOne (see Figure~\ref{fig:ROI_ZUR}) and apply the trained models without fine-tuning in \zurichThree. Note the sharp crease building edges and the reconstructed roof details shown in the zoomed oblique view (orange rectangle).}
    \label{fig:generalization_ZUR3}
\end{figure*}
We summarize quantitative results in Table~\ref{tab:generalization_ZUR} and show visual examples in Figure~\ref{fig:generalization_ZUR2} and Figure~\ref{fig:generalization_ZUR3}. \mbox{\ours}-stereo performs exceptionally well when generalizing over larger distances. The model is able to reconstruct sharp building outlines and roof features (see Figure~\ref{fig:generalization_ZUR2}(c) and Figure~\ref{fig:generalization_ZUR3}(c), zoomed oblique view) and even recovers reasonable terrain heights under a closed canopy (see Figure~\ref{fig:generalization_ZUR3}(c), upper part of the scene). Moreover, it separates built-up areas into plausible individual units, even for challenging urban structures such as the historic city center (see Figure~\ref{fig:generalization_ZUR2}(c), right). Still, in some cases, buildings are reproduced inaccurately or reconstructed only partially. For \zurichTwo, \ours-stereo yields an overall MAE of 2.0$\,$m and a MedAE of 1.0$\,$m. For \zurichThree, the overall MAE is 2.6$\,$m and the MedAE 2.0$\,$m. Compared to the conventional stereo pipeline, \ours-stereo reduces the overall MAE by up to 48\% and the MAE of buildings by up to 30\%.

Quantitatively, \ours-stereo and \resdepth-stereo perform very similarly on the test region in \zurichOne, with almost no domain shift between the training and test sites. When generalizing from \zurichOne to \zurichTwo or \zurichThree, \ours-stereo consistently outperforms \resdepth-stereo across all error metrics. The overall MAE is up to 18\% lower, with the largest accuracy gain on buildings. Visually, \ours-stereo produces smoother terrain, reconstructs straighter and crisper surface creases, and recovers more detailed roof geometry.

\end{document}